\documentclass[nodoi, nokeyword, nocopyright, nocopysymb]{nle}

\usepackage{amsmath}
\usepackage{graphicx}
\usepackage{natbib}
\ifpdf%
\usepackage{epstopdf}%
\else%
\fi

\usepackage{url}
\usepackage{xspace}

\usepackage{tikz}

\usepackage{drs}

\usepackage{amsmath}
\usepackage{tikz-dependency}
\usepackage{gb4e}
\usepackage{mwe}
\usepackage{multirow}

\usepackage{tikz-qtree}

\tikzset{baseline,every tree node/.style={align=center,anchor=north}}

\usepackage[normalem]{ulem}

\begin{document}
\label{firstpage}


\papertitle{Article}

\jnlPage{1}{00}
\jnlDoiYr{2019}
\doival{10.1017/xxxxx}

\newcommand{\eat}{EAT\xspace}
\newcommand{\eatseq}{EAT2seq\xspace}
\newcommand{\eatsnlg}{EAT-SimpleNLG\xspace}

\title{\eat: a simple and versatile semantic representation format for multi-purpose NLP}

\author{Tommi Gr\"{o}ndahl}
\affiliation{Aalto University, Konemiehentie 2, 02150 Espoo, Finland,
\email{tommi.grondahl@aalto.fi}}

\begin{abstract}
\emph{Semantic representations} are central in many NLP tasks that require human-interpretable data.
The \emph{conjunctivist} theoretical framework -- primarily developed by \citet{Pietroski05a, Pietroski18} -- obtains expressive representations with only a few basic semantic types and relations systematically linked to syntactic positions.
While representational simplicity is crucial for computational applications, such findings have not yet had major influence on NLP.
We present \emph{the first generic semantic representation format for NLP} directly based on these insights.
We name the format \emph{\eat} due to its basis in the Event-, Agent-, and Theme arguments in Neo-Davidsonian logical forms. It builds on the idea that similar tripartite argument relations are ubiquitous across categories,
and can be constructed from grammatical structure without additional lexical information.
We present a detailed exposition of \eat and how it relates to other prevalent formats used in prior work, such as Abstract Meaning Representation (AMR) and Minimal Recursion Semantics (MRS).
\eat stands out in two respects: \emph{simplicity} and \emph{versatility}.
Uniquely, \eat discards semantic metapredicates, and instead represents semantic roles entirely via \emph{positional encoding}. This is made possible by limiting the number of roles to only \emph{three}; a major decrease from the many dozens recognized in e.g. AMR and MRS.
\eat's simplicity makes it exceptionally versatile in application.
First, we show that drastically reducing semantic roles based on \eat benefits text generation from MRS in the test settings of \citet{Hajdiketal2019}.
Second, we implement the derivation of \eat from a syntactic parse, and apply this for \emph{parallel corpus generation} between grammatical classes.
Third, we train an encoder-decoder LSTM network to map \eat to English.
Finally, we use both the encoder-decoder network and a rule-based alternative to conduct \emph{grammatical transformation} from \eat-input.
Our experiments illustrate \eat's ability to retain semantic information despite its simplicity.
\end{abstract}

\maketitle

\section{Introduction}

\emph{Semantic representations} are needed in NLP tasks that abstract away from surface-level grammar and require human-interpretability.
Traditional formal semantics relies on ideas formulated by \citet{Frege1879}, and applied to natural language by \citet{Montague70, Montague73}. Important advances have been made within this framework, but it has certain difficulties for computational implementation. It requires \emph{fixed valencies} for lexical items; relies on elaborate semantic types that increase in complexity along with linguistic coverage; and places no \emph{a priori} restrictions on possible semantic types.
One way to mitigate such problems is to restrict possible semantic types to as few as possible.
This is the basis of Paul Pietroski's \emph{conjunctivist} project, so far developed in two monographs and many related papers \citep{Pietroski03, Pietroski05a, Pietroski18, Pietroski10, Pietroski11, Hornstein:Pietroski09}.
Conjunctivism is essentially an extension of \emph{Neo-Davidsonian} semantics \citep{Higginbotham85, Parsons90, Schein93} in a broader scale.

Prior work in semantic parsing has used Neo-Davidsonian argument structures \citep{Bos2015, Reddyetal2016, Reddyetal2017}, and conjunctivist ideas have occasionally been used in computational implementations for specific syntactic frameworks \citep{Gaudreault2016}.
However, the theoretical insights have not yet made their way to general NLP applications with broad data coverage.
This paper presents the first comprehensive semantic representation format that is both (i) explicitly built on the conjunctivist framework, and (ii) specifically tailored for NLP, with wide applicability to \emph{information retrieval} and \emph{text generation tasks}.

We call the format \textit{\eat} due to its connection with the Event, Agent and Theme arguments in \mbox{Neo-Davidsonian} semantics (Section \ref{sec:EAT}).
\eat can be built directly from a syntactic parse without the use of external lexical knowledge bases.
It uses positional encoding for argument relations (Section \ref{sec:EAT-definition}), uninflected lemmas to represent words, and Boolean features for grammatical properties (Section \ref{sec:grammatical-features}).
Multiple \eat-tuples are presented in sequence with a conjunctive default interpretation.
This provides a concrete implementation of the basic conjunctivist theory (Section \ref{sec:conjunctivism}).
We also discuss extensions needed to apply conjunctivism across more complex semantic phenomena (Section \ref{sec:conjunctivism-refinements}), and how these could be incorporated to \eat (Section \ref{sec:EAT-refinements}).

We argue that \eat is beneficial on two main grounds compared to alternative semantic representation frameworks: \emph{simplicity} and \emph{versatility} (Section \ref{sec:EAT-comparison}).
\eat only uses \emph{three} semantic roles along with certain additional features that regulate their interpretation in different constructions.
Since there are so few roles, it is possible to represent them by assigning each to a dedicated position in a word triplet.
This positional encoding is not practically feasible with prior formats that use dozens of roles \citep{Banarescuetal13, Copestakeetal2005}.
\eat's versatility manifests in at least four main ways.
First, its simple and interpretable structure makes it easy to navigate in information retrieval tasks.
Second, it can be derived from a standard syntactic parse with no recourse to external lexical information (Section \ref{sec:EAT-from-parse}).
Third, its linearity makes it directly vectorizable.
Finally, its structure allows very easy modification.

We applied of \eat across multiple tasks on English data (Section \ref{sec:Results}).
The first of these involves significantly reducing argument roles from another format -- Minimal Recursion Semantics (MRS) \citep{Copestakeetal2005} -- in the text generation experiments of \citet{Hajdiketal2019}. Our results illustrate the benefit of such reduction on model performance (Section \ref{sec:comparison-mrs}).
For showing the ability of \eat to retain and modulate information in the syntactic parse, we generated \emph{parallel corpora} between grammatical classes (Section \ref{sec:parallel-corpora}).
Finally, we used an encoder-decoder LSTM network and the \emph{SimpleNLG} surface realizer \citep{Gatt:Reiter2009} for generating English from \eat and applying \emph{grammatical transformation} (Section \ref{sec:EAT2seq}).
In addition to outlining the usefulness of \eat as a simple but powerful representation format, our experiments yield multiple novel parallel corpora for the NLP community.
The list below summarizes our contributions.

\begin{itemize}
\item We present \eat: a novel, highly simple but expressive semantic representation format motivated by the conjunctivist framework in semantic theory (Sections \ref{sec:Background}--\ref{sec:EAT}).
\item We implement \eat-construction from two syntactic parsing frameworks: \emph{probabilistic context-free grammar} (Section \ref{sec:PCFG}) and \emph{dependency grammar} (Section \ref{sec:dependency-EAT}).
\item We show that a drastic reduction of semantic roles (motivated by \eat) benefits text reconstruction from MRS (Section \ref{sec:comparison-mrs}).
\item We use \eat to generate multiple parallel corpora between grammatical classes, and provide these corpora as open-access resources (Sections \ref{sec:parallel-corpora}; \ref{sec:EAT2seq}).\footnote{We provide our source code, the trained encoder-decoder network, and the parallel corpora here: \url{https://drive.google.com/drive/folders/1_kqvFQiwiurkZA39By3xNiXPhjhMCLrG?usp=sharing}. If the paper is accepted for publication, we will make them available on GitHub.}
\item We use both an encoder-decoder network and SimpleNLG to generate English from \eat-input, and apply this to grammatical transformation (Section \ref{sec:EAT2seq}).
\item We apply \eat for effective automatic self-assessment of the generated output (Section \ref{sec:EAT2seq}).
\end{itemize}

\section{Background}
\label{sec:Background}

In this section we discuss the theoretical background in terms of semantics (Section \ref{sec:argument-structure-semantics}), syntax (Section \ref{sec:argument-structure-syntax}), and their interface (Section \ref{sec:conjunctivism}).
\eat is motivated by two main ideas: (i) reducing the semantic types and operations to the bare minimum; and (ii) a systematic mapping between syntactic and semantic structure \emph{without} a separate lexical knowledge base.
The work of \citet{Pietroski05a, Pietroski11, Pietroski18} has combined these desiderata in the most elaborative manner to date.
His \emph{conjunctivist} analysis functions as the main motivation for \eat.

The linguistic theory behind conjunctivism mostly stems from the \emph{generative} framework. We present a brief exposition of the syntactic background in Section \ref{sec:argument-structure-syntax}, in order to better explain the choices we later make when mapping a syntactic parse to \eat (Sections \ref{sec:PCFG}; \ref{sec:dependency-EAT}). However, the \eat format could also be connected to alternative linguistic theories like cognitive grammar \citep{Langacker87, Lakoff87} or construction grammar \citep{Croft01, Goldberg06}. We leave such considerations for future work.

%
%
\subsection{Argument structure in semantics}
\label{sec:argument-structure-semantics}

The classical Fregean analysis of verbs treats them as predicates that take thematic arguments \citep{Frege1879}.
\citet{Montague70, Montague73} presented a general strategy for Fregean formalization of natural languages based on techniques developed by \citet{Church36} for expressing functions of arbitrary complexity.
The fundamental ontological types in this system are \emph{entities} ($e$) and \emph{truth-values} ($t$), and all other types are functions from some type to another. A function from type $a$ to type $b$ is notated as \mbox{$<a,b>$}.\footnote{We abstract away from lambda-notation for simplicity. Linguistically oriented introductions to Montague semantics are provided in \citet{Parteeetal90} and \citet{Heim:Kratzer98}.}
Verb \emph{valency} is thus determined by its type. An \emph{intransitive} verb is of the type \mbox{$<e,t>$}: a function from an entity to a truth-value. A \emph{transitive} verb is of the type \mbox{$<e,<e,t>>$}: a function from an entity to another function from an entity to a truth-value.
Example (\ref{LF1}) shows the semantic derivation of an example intransitive and transitive clause.

\begin{exe}
\ex \label{LF1}
\begin{minipage}[t]{0.35\textwidth}
\Tree [.{John runs: $<t>$} {John: $<e>$} {runs: $<e,t>$} ]
\end{minipage}
\begin{minipage}[t]{0.5\textwidth}
\Tree [.{John sees Mary: $<t>$} {John: $<e>$} [.{sees Mary: $<e,t>$} {sees: $<e,<e,t>>$} {Mary: $<e>$} ]]
\end{minipage}
\end{exe}

In simpler notation, intransitive verbs are \emph{monadic predicates}, and transitive verbs are \emph{dyadic predicates}. Using capital letters for semantic content, the analysis of \emph{John runs} is \mbox{RUN(JOHN)}, and the analysis of \emph{John sees Mary} is \mbox{SEE(JOHN, MARY)}. Verb valency is assimilated to \emph{predicate adicity}: the number of ``slots'' that need to be filled to arrive at a truth-value.
%
%
Davidson later introduced \textit{event variables} \citep{Davidson67, Castaneda67}, which he argued to be present in all verbs and existentially quantified in (declarative) sentences. The intuition is that the sentence affirms or negates the existence of the kind of event described by the main verb.
%
%
Event variables allow a straight-forward account of certain verb modifiers like (subsective)\footnote{See Section \ref{sec:conjunctivism} for discussion of non-subsective modifiers.} adverbs, which are additional predicates over the event variable.
%

While Davidson's original analysis adds event variables to the picture, it does not otherwise alter the Fregean system.
In particular, verbs still have an inherent semantic valency.
An alternative was presented within the \textit{Neo-Davidsonian} framework, where thematic arguments are removed from the verb, and each thematic role is allocated to a separate dyadic relation between the event variable and a thematic argument \citep{Higginbotham85, Parsons90, Schein93, Pietroski05a, Lohndal14}. We use the standard terms Agent and Theme for the two arguments of a transitive verb.\footnote{\emph{Patient} is another common term to use instead of Theme. We make no notational distinction between these two here, as we take this difference to require further information about the verb and/or argument in question (see e.g. \citealt{Jackendoff87, VanValin99}), and hence not to be decidable based on syntax alone.}
%
%
The Neo-Davidsonian analysis assigns all verbs to the \emph{same} simple semantic type P($e$): a monadic predication over an event variable (type \mbox{$<e,t>$} in the Montagovian formalism).
One of its major benefits is allowing a simple treatment of \emph{transitivity alterations}, where the same verb can appear with or without a thematic argument.
For example, the Neo-Davidsonian analysis of \emph{John bakes bread} is \mbox{$\exists e \exists x$ [BAKE($e$) $\land$ Agent($e$, JOHN) $\land$ BREAD(x) $\land$ Theme($e$, $x$)]}, and accounting for the intransitive \emph{John bakes} is trivial simply by removing the Theme.

\emph{Prepositions} function as dyadic relations over two elements, and are thus importantly similar to Neo-Davidsonian thematic roles \citep{Hale:Keyser02}.
For prepositions denoting concrete relations, the first argument can often be assigned to the so-called \emph{Figure}, and the second argument to the \emph{Ground} \citep{Talmy78, Svenonius08}.
%
%
%
%
\emph{Connectives} relate two clauses. In simple cases they can be analyzed as connecting two events.
However, many connectives do not allow a conjunctive treatment, as they do not entail one or either of the clauses they relate.
%
%
Similar considerations also apply to certain other types of grammatical elements, and their inclusion complicates the analysis. As its name suggests, the \emph{conjunctivist} analysis (Section \ref{sec:conjunctivism}) is based around the simple conjunctive cases, and should be extended only after it has first been shown to work with them.
We discuss non-conjunctive modifiers and arguments in Section \ref{sec:conjunctivism-refinements}.


\subsection{Argument structure in syntax}
\label{sec:argument-structure-syntax}

We provide a brief exposition of phrase structure building and the syntax-semantics interface, remaining broadly within the current mainstream generative framework \citep{Newmeyer1996, Lasnik:Lohndal13}, since this is an explicit component of conjunctivism as developed by \citet{Pietroski05a, Pietroski11, Pietroski18} (Section \ref{sec:conjunctivism}). As future work, the application of similar ideas within alternative linguistic frameworks would also be beneficial.
%

Early generative grammars were construed as language-specific sets of \emph{phrase structure rules} \citep{Chomsky57, Chomsky65}.
%
%
These were later replaced with \emph{X'-theory}, which unifies all phrases to a generic format \citep{Chomsky70, Chomsky81, Jackendoff77, Haegemann94}.
Here, each \emph{phrase} is projected from a syntactically atomic \emph{head}, and can contain three types of phrases as arguments or modifiers: \emph{complements}, \emph{specifiers}, and \emph{adjuncts}.
Complements are combined with the head to form an intermediate \emph{X'}-phrase.
This is then combined with a specifier to produce the maximal phrase \emph{XP}.
Adjuncts are optional modifiers of X' that recursively project another instance of it.
Phrase-structure rules and notation for the X'-schema are provided in (\ref{XBAR}). \\

\begin{exe}
\ex \label{XBAR}
\begin{minipage}[t]{0.3\textwidth}
\begin{xlist}
\ex XP $\rightarrow$ YP X'
\ex X' $\rightarrow$ WP X'
\ex X' $\rightarrow$ X ZP
\end{xlist}
\end{minipage}
\begin{minipage}[t]{0.45\textwidth}
\Tree [.XP {YP\\(Spec,XP)} [.X' WP\\(adjunct) [.X' X\\(head) ZP\\(Comp,XP) ]]]
\end{minipage}
\end{exe}

By restricting the range of possible phrase-structural relations, X'-theory made possible the linking between syntactic and semantic representations much more systematically than before.
\citet{Baker88} famously suggested that the thematic roles of Agent and Theme are in a deterministic correspondence with the X'-schema in verb phrases (VPs). This principle is known as the \emph{Uniformity of Theta Assignment Hypothesis} (UTAH). A common variant of UTAH allocates Agent to Spec,VP and Theme to Comp,VP.
\emph{Clauses} are built on top of VPs with additional \emph{functional heads}.
These behave similarly to lexical heads (like verbs and nouns) in terms of X'-structure, but serve a grammatical rather than lexical purpose \citep{Chomsky86, Abney87, Cinque:Rizzi16}.
VP is the complement of a functional head called \emph{T} (or sometimes \emph{I}) that hosts inflectional material like tense or mood \citep{Chomsky81, Pollock89, Ritter:Wiltschko14}.
TP is the complement of a functional head called \emph{C} that hosts \emph{complementizers} (e.g. \emph{that}) and determines discourse-features like as clause type (declarative, question, command), topicalization, and focus \citep{Rizzi97}.

In contrast to the thematic interpretation of verbal arguments via UTAH, functional specifier positions are typically landing sites for \emph{movement}, i.e. the re-introduction of an element higher in the structure.
Spec,TP most closely corresponds to the canonical ``subject'' position in English. The Agent raises from Spec,VP to Spec,TP in active clauses; and the Theme raises from Comp,VP to Spec,TP in passive clauses.
Spec,CP hosts focused elements, and most notably \emph{wh}-question phrases raised from original positions determining their thematic role.
The C-T-V clause structure is shown in (\ref{CTV}), along with typical interpretations for each position.

\begin{exe}
\ex \label{CTV}
\begin{tikzpicture}
\Tree [.CP Spec,CP\\{(\emph{wh}-phrases;}\\{focus)} [.C' C\\{(clause type)} [.TP Spec,TP\\(subject) [.T' T\\(tense) [.VP Spec,VP\\(Agent) [.V' V\\(Event) Comp,VP\\(Theme) ]]]]]]
\end{tikzpicture}
\end{exe}

In subsequent research, C and T have been expanded to many more functional heads (\citealt{Rizzi97, Cinque99, Belletti04, Kayne05, Cinque:Rizzi10}; \citeyear{Cinque:Rizzi16}).
A common strategy has been to treat C and T as shorthands for \emph{domains} that can include multiple functional heads \citep{Grohmann03, vanGelderen13, Ramchand:Svenonius2014, Wiltschko14}.
VP-internal material is responsible for \emph{argument structure}, the T-domain specifies various aspects of verb \emph{inflection}, and the C-domain assigns \emph{discourse-level} semantics.

\emph{Nouns} can project arguments and functional structure in a similar manner as verbs \citep{Alexiadouetal07}.
\emph{Possessors} are usually taken to occupy the specifier position of either the noun itself \citep{Chomsky70, Jackendoff77} or nominal functional projection \citep{Abney87, Adger03}. Nouns typically appear in argument positions of either a verb, another noun, or a preposition.
In classical X'-theory, \emph{adjectives} and \emph{adverbs} are prototypical \emph{adjuncts} of nouns or verbs, respectively (e.g. \citealt{Jackendoff77}).
They project additional functional structure expressing \emph{comparison class} \citep{Bobaljik2012}, and can be further modified by \emph{degree adverbs} (e.g. \emph{very}).

\emph{Prepositions} denote dyadic relations (Section \ref{sec:argument-structure-semantics}).
The internal argument is the complement of P, but there is more controversy concerning the syntactic status of the external argument.
Some analyses treat the PP as only containing the P and its complement (Ground), and being combined with the Figure as an adjunct or a complement (e.g. \citealt{Hornstein:Pietroski09}). Another option is to assign the arguments to complement and specifier positions inside the PP \citep{Hale:Keyser02}.
Despite their differences, both analyses manifest a similar abstract schema, where the Ground is the complement of P, and the Figure is a higher element adjacent to the preposition-complement construction.
Figure (\ref{PP}) shows the alternative analyses on left and center (non-branching X'-positions omitted), and the generic scheme on the right.

\begin{exe}
\ex \label{PP}
\begin{minipage}[t]{0.3\textwidth}
\Tree [.XP X\\(Figure) [.PP P Comp,PP\\(Ground) ]]
\end{minipage}
\begin{minipage}[t]{0.3\textwidth}
\Tree [.PP Spec,PP\\(Figure) [.P' P Comp,PP\\(Ground) ]]
\end{minipage}
\begin{minipage}[t]{0.3\textwidth}
\Tree [.{} Figure [.{} P Ground ]]
\end{minipage}
\end{exe}


The treatment of \emph{connectives} mirrors that of prepositions. These combine two elements into a complex element of the same syntactic type. Clausal connectives can often be allocated to the C-domain in the clause.
Like with prepositions above, we remain agnostic about the specific placement of the higher clause, but take the lower to be the complement of the connective.
\emph{Coordinating} heads (e.g. \emph{and, or}) have also been argued to form their own X'-theoretic projections, with the coordinated elements in complement and specifier positions \citep{Kayne94, Johanessen98}.


The generic phrase-structure scheme (\ref{PS-schema}) thus emerges across the different categories discussed in this section, where a \emph{head} first takes a more proximal \emph{internal argument}, and subsequently a higher \emph{external argument}.

\begin{exe}
\ex \label{PS-schema}
\Tree [.{} External\\argument [.{} Head Internal\\argument ]]
\end{exe}

The scheme (\ref{PS-schema}) resembles X'-theory but can depart from it.
For example, it is common to treat the Agent argument as introduced by a dedicated functional projection instead of the verb itself \citep{Kratzer96}, and some analyses have extended this to Theme as well \citep{Borer05b, Lohndal14}. However, even here the Agent is a higher argument combined with a verb-Theme complex lower in the structure.
Also, as discussed above, we can abstract away from whether the external argument of a preposition is assigned at Spec,PP or is a separate phrase taking the PP as an adjunct.
We will use (\ref{PS-schema}) instead of strict X'-theory, we will use the notation \emph{Int,X} and \emph{Ext,X} for the internal and external arguments of the head X.

\subsection{Conjunctivism: a minimal but expressive syntax-semantics interface}
\label{sec:conjunctivism}

The syntax-semantics interface must support \emph{compositionality}, where the meaning of a complex expression is determined by the meanings of its constituent expressions and their mode of combination \citep{Frege1879}.
As discussed in Section \ref{sec:argument-structure-semantics}, the standard approach in formal semantics is to achieve compositionality via \emph{function application} \citep{Montague70, Montague73, Heim:Kratzer98}, where one element is a function that takes the other as an argument.
As an alternative, Pietroski's (\citeyear{Pietroski05a, Pietroski18}) account builds on \emph{conjunction}.
We go through the basics of the conjunctivist system here, and discuss its extension to more challenging cases in Section \ref{sec:conjunctivism-refinements}.
%

A compositional theory of the syntax-semantics interface should account for the semantics of syntactic atoms (with no further decomposition), and the semantics of each syntactic mode of combination.
In Section \ref{sec:argument-structure-syntax} we specified the nature of syntactic system to a sufficient degree for present purposes.
Atomic syntactic elements are \emph{heads} of two possible types: \emph{lexical heads} that project phrases, and \emph{functional heads} that refine the phrases by introducing grammatical features and landing sites for movement. Syntactic modes of combination come in (at least) four kinds: (i) those that follow the argument-introducing scheme (\ref{PS-schema}), (ii) adjunction, (iii) adding functional heads, and (iv) movement.
We abstract away from semantic aspects of movement (related to e.g. scope or focus), and interpret elements in their pre-movement positions. This leaves syntactic heads (Section \ref{sec:conjunctivism-heads}) and the three remaining types of syntactic combination (Section \ref{sec:conjunctivism-phrases}).

\subsubsection{Semantics of syntactic heads}
\label{sec:conjunctivism-heads}

Starting with lexical heads, \emph{nouns} are monadic predicates (over e.g. objects, masses, or events).
The traditional analysis of \emph{proper names} as logical constants has also been challenged in favour of a predicative approach \citep{Quine60, Burge1973, Pietroski05a, Pietroski18}.\footnote{The two main ways of treating a proper name as a predicate are to assimilate it to some definite description \citep{Quine39}, or to equivocate a predicate P($x$) with the identity statement $x=c$, where $c$ is a constant \citep{Quine48, Quine60}. The status of names as predicates is further corroborated by the linguistic observation that in many languages they behave like common nouns, e.g. requiring articles \citep{Anderson2007}.}
The Neo-Davidsonian framework further allows treating all \emph{verbs} as monadic predicates over an event variable, irrespective of transitivity (Section \ref{sec:argument-structure-semantics}).
Finally, discarding non-subsective variants for now (see Section \ref{sec:conjunctivism-refinements}), \emph{adjectives} can be assimilated to nouns and \emph{adverbs} to verbs in these respects.
Hence, we can unify the semantic interpretation of all four major types of lexical heads as monadic predicates.

The status of \emph{prepositions} as lexical or functional heads is controversial, along with whether they should be thought of as dyadic relations or monadic predicates of \emph{states} akin to events (e.g. \citealt{Svenonius08, Pietroski18}).
Following Section \ref{sec:argument-structure-semantics}, we adopt the relational analysis here, and extend it to those \emph{connectives} that entail what they connect (e.g. \emph{and}, \emph{but}, \emph{while}, \emph{whereas}).
For simplicity, we call both ``prepositions'' in the remainder of this section, and leave the discussion of non-entailing connectives to Section \ref{sec:conjunctivism-refinements}.
Prepositions expand the range of semantic types to include \emph{dyadic relations}, but these remain restricted to a closed class.

Functional heads extend the phrase with additional grammatical information. Many of these have semantic types that go beyond first-order predication (e.g. negation or modality), which we review in Section \ref{sec:conjunctivism-refinements}. However, some allow a simpler analysis. For example, if \emph{plural variables} are used \citep{Boolos84, Schein93, Pietroski03, Pietroski05a, Pietroski18, Lohndal14}, \emph{number} in nouns can be treated as a predicate over such a variable, indicating how many values it has. We can thus assimilate some functional heads to other modifiers in semantic type as monadic predicates.

In summary, we have outlined a very austere position where each syntactic head is a \emph{monadic predicate}, apart from the closed class of prepositions that are \emph{dyadic relations}.
Whatever other differences exist between the semantics of different lexical heads, they do not concern the basic semantic type.
This restriction also allows limiting semantic combination accordingly.

\subsubsection{Semantics of complex phrases}
\label{sec:conjunctivism-phrases}

In simple cases, we assimilated nouns, verbs, adjectives, adverbs, and their respective functional heads as monadic predicates in semantic type.
We begin with the most basic combination of two such elements.
The first of such methods is \emph{adjunction}, and the second is adding a functional head to the phrase projected from a lexical head.
As shown in (\ref{conj-conjunction}), in both variants the two predicates are applied over the same (free) variable, and these predications are \emph{conjoined}.

\begin{exe}
\ex \label{conj-conjunction}
\Tree [.{P($x$) $\land$ Q($x$)} P($x$) Q($x$) ]
\end{exe}

The basic conjunction rule (\ref{conj-conjunction}) is the \emph{default} operation applied for the syntactic combination of two elements, and serves as the backbone of conjunctivism \citep{Pietroski05a, Pietroski18}. It is applied when no other rule is available based on stricter syntactic criteria.
Neo-Davidsonian event semantics allows the same analysis for modifiers of both noun and verb phrases.

%
%
Conjunction alone obviously cannot bring about argument structure.
In the spirit of UTAH \citep{Baker88}, Pietroski argues that dedicated syntactic positions map to thematic roles.
Following the discussion in Section \ref{sec:argument-structure-syntax}, the Theme position is Int,V and the Agent position Ext,V, shown in (\ref{conj-UTAH}).
The event variable linked to the verb remains free, while introducing a thematic argument existentially closes the variable linked to it.
The arguments are introduced via conjunction.
(Note that either argument can also occur in the absence of the other, as in intransitives or passives.)

\begin{exe}
\ex \label{conj-UTAH}
\begin{minipage}[t]{0.2\textwidth}
\Tree [.{} Ext,V [.{} V Int,V ]]
\end{minipage}
\begin{minipage}[t]{0.4\textwidth}
\Tree [.{P($e$) $\land$ $\exists x$[Q($x$) $\land$ Agent($e$, $x$)] $\land$ $\exists y$[W($y$) $\land$ Theme($e$, $y$)]} {Q($x$)} [.{P($e$) $\land$ $\exists y$[W($y$) $\land$ Theme($e$, $y$)]} {P($e$)} {W($y$)} ]]
\end{minipage}
\end{exe}


PPs denote dyadic relations (Sections \ref{sec:argument-structure-semantics}; \ref{sec:conjunctivism-heads}), and follow the syntactic scheme (\ref{PS-schema}) with Int,P as the first relatum and Ext,P as the second \citep{Hornstein:Pietroski09}.
As shown in (\ref{conj-PP}), Int,P and Ext,P are monadic predicates, and Int,P is existentially closed while Ext,P remains free. Hence, e.g. \emph{a dog in a house} is true of some dogs rather than some houses.

\begin{exe}
\ex \label{conj-PP}
\begin{minipage}[t]{0.40\textwidth}
\Tree [.{PP} Ext,P [.{} P Int,P ]]
\end{minipage}
\begin{minipage}[t]{0.5\textwidth}
\Tree [.{P($x$) $\land$ $\exists y$[Q($y$) $\land$ R($x$, $y$)]} {P($x$)} [.{$\exists y$[Q($y$) $\land$ R($x$, $y$)]} {R($x$, $y$)} {Q($y$)} ]]
\end{minipage}
\end{exe}

The conjunctivist system easily allows adding further thematic roles for \emph{indirect objects}.
Syntactically, they have been argued to occupy dedicated positions within the extended VP, akin to Theme and Agent \citep{Ramchand08, Pylkkanen2008}.
Semantically, they function like PP-modifiers, which is in line with the complementarity of the \emph{to}-PP with the dative indirect object in English. For present purposes we assimilate these in semantics, equivocating e.g. \emph{Mary gave John a flower} with \emph{Mary gave a flower to John}. (Note that this does not mean either construction is syntactically derived from the other; only that they share roughly the same interpretation.)

We now have the first version of the basic conjunctivist system. Thematic arguments of verbs are assigned as in (\ref{conj-UTAH}), prepositions and their relata as in (\ref{conj-PP}), and other combinations (adjunction or adding functional heads) via the default rule (\ref{conj-conjunction}).
Importantly, every \emph{complex phrase} in (\ref{conj-conjunction})--(\ref{conj-PP}) is \emph{monadic}, containing one free variable. This is achieved by existentially quantifying over other variables as specified in (\ref{conj-UTAH})--(\ref{conj-PP}). The determination of which element the free variable is linked to depends on the construction: the verb head in VPs/clauses (event variable), and Ext,P in PPs.

\subsection{Refinements to conjunctivism}
\label{sec:conjunctivism-refinements}

The rules in Section \ref{sec:conjunctivism} account for \emph{conjunctive} semantic combination, where the contents of constituent expressions are entailed by the content of the full expression.
However, certain more complex aspects of semantics resist such treatment. Thorough analysis of these quickly becomes complex, and here we will only focus on the most fundamental question for present purposes: what \emph{semantic types} and \emph{modes of combination} are needed beyond monadic and dyadic predicates and conjunction?
We argue that while the basic scheme requires some significant additions, these can still be limited to only few in kind and number.

\subsubsection{Negation, modality, and non-subsective modifiers}

Starting with simple but crucial case, \emph{negation} is evidently non-conjunctive.
It must be some kind of an \emph{operator} that takes the meaning of the negated clause as an argument.
The standard Fregean analysis is that it switches a \emph{truth-value} from \emph{true} to \emph{false} or vice versa; hence having the type \mbox{$<t,t>$}.
However, \citet{Pietroski18} presents an alternative based on \citet{Tarski1944}, which treats negation as a \emph{predicate modifier} of the type \mbox{$<<e,t>, <e,t>>$} instead.

Two ``arrow'' operators can be defined as follows: \mbox{$\Uparrow$P($x$)} applies to everything if P($x$) applies to something (i.e. \mbox{$\exists y$P($y$)}) and otherwise applies to nothing; and \mbox{$\Downarrow$P($x$)} applies to everything if \mbox{P($x$)} applies to nothing (i.e. \mbox{$\neg\exists y$P($y$)}) and otherwise applies to nothing.
Pietroski assigns both to the T-domain in English clausal syntax (see Section \ref{sec:argument-structure-syntax}), where negation corresponds to the downward arrow.
Simply put, $\Downarrow$P($e$) can be defined as `$x$ is such that $\neg \exists e $P($e$)'.
This technical maneuver allows maintaining the conjunctivist tenet of every complex phrase being a monadic predicate.

\emph{Modal operators} applied to P($x$) can be analyzed as `$x$ is such that it is possible/necessary/(...) that P($x$)'.
Understood like this, they are also predicate modifiers.
The modal statement inside the quotes can then be opened further by using more elaborate approaches that involve e.g. possible world semantics \citep{Lewis1986, Kratzer2012}; but this does not alter the semantic type itself.
Significantly, the same analysis can be applied to different modal elements regardless of their syntactic status as an auxiliary, adverb, or adjective.
This shows one major benefit of the arrow operator analysis of negation: $\Downarrow$P($x$) is of the same type as P($x$), and hence the same modal operators can apply to either without further complications.
Two examples are shown in (\ref{Conj-modals}).

\begin{exe}
\ex \label{Conj-modals}
\begin{xlist}

\ex a possible thief \hfill $\Leftrightarrow$ \\
POSSIBLE[THIEF($x$)] \hfill $\Leftrightarrow$ \\
$x$ is such that it is possible that THIEF($x$)

\ex It might not rain \hfill $\Leftrightarrow$ \\
MIGHT[$\Downarrow$RAIN(e)] \hfill $\Leftrightarrow$ \\
$x$ is such that it is possible that [$x$ is such that $\neg\exists e$RAIN($e$)]

\end{xlist}
\end{exe}

The analysis of the modal adjective in (\ref{Conj-modals}a) can be extended further to other types of \emph{non-subsective} adjectives/adverbs, such as \emph{fake}, \emph{former(ly)}, or \emph{alleged(ly)}. Abstracting away from the precise analysis of their lexical content, in semantic type they are predicate modifiers (\mbox{$<<e,t>, <e,t>>$}).
Additionally, \emph{degree adverbials} (e.g. \emph{very}) are predicate modifiers of a similar kind, and take adjectives or adverbs as syntactic arguments.
%
%
%
%
%

Thus, many non-entailing elements can be unified in semantic type by analysing them as predicate modifiers.
The combination mechanism is not conjunctive, but instantiates \emph{function application}:
predicate modifiers are \emph{second-order functions} from predicates to others.
This recourse to function application might initially seem to break away from the original idea behind conjunctivism as an alternative to the Montagovian framework. However, its use here is still much more limited than in the standard system.

\subsubsection{Connectives and quantifiers}

In addition to predicate modifiers, we need to account for non-entailing \emph{relations} such as \emph{connectives}.
The standard Fregean/Montagovian analysis is to treat them as functions from two truth-values to a third: \mbox{$<t, <t,t>>$}. In line with prior discussion, we can replace this with \mbox{$<<e,t>, <<e,t>, <e,t>>>$}: a function from two monadic predicates to a third.
For example, the connective \emph{or} can be analyzed as a function from two predicates to a third that applies to $x$ if $x$ applies to either of the argument predicates.
%

The affinity between connectives and \emph{quantifiers} (e.g. \emph{all}, \emph{some}) is central in some theoretical frameworks, such as Discourse Representation Theory \citep{Kamp81, Kamp95}.
Adopting plural logic (see Section \ref{sec:conjunctivism-heads}) also allows some quantifiers to receive a simpler analysis as plural predicates akin to number (e.g. \emph{many}, \emph{few}).
\citet{Pietroski03, Pietroski05a, Pietroski18} discusses quantifiers extensively within the conjunctivist framework and provides an alternative account, but we abstract away from this due to its technical complexity.\footnote{Pietroski's analysis has certain theoretical benefits, in particular accounting for the so-called \emph{conservativity} of natural language quantifiers, which is a central semantic property they share. The main cost of his account is the requirement of a novel semantic type not present elsewhere ($<t,e>$).}
Here, it suffices that the semantic type \mbox{$<<e,t>, <<e,t>, <e,t>>>$} is needed for connectives, and could be extended to quantifiers.

\subsubsection{Clausal arguments}
\emph{Propositional attitude ascriptions} take clausal arguments that do not denote events but the contents of the clauses themselves.
Philosophical literature includes many alternative suggestions of what such contents could be, in terms of e.g., \emph{intensional logic} \citep{Montague73} or \emph {interpreted logical forms} \citep{Larson:Ludlow1993}.
Without going further into this discussion here, it suffices to follow \citet{Pietroski00, Pietroski05a} in taking some verbs to require \emph{Content} arguments instead of a Theme (e.g. \emph{think}, \emph{believe}).
Both appear in the same syntactic position (Int,V).
This addition complicates the the conjunctivist scheme by being \emph{intensional}, as the Content argument is the clause's meaning itself and not the event variable associated with it.

\subsubsection{Summary of refinements to conjunctivism}
In this section we have reviewed three types of additions to the basic conjunctivist system presented in Section \ref{sec:conjunctivism}: \emph{predicate modifiers}, \emph{connectives}(/\emph{quantifiers}), and \emph{Content arguments}.
Predicate modifiers have the type $<<e,t>, <e,t>>$, and account for \emph{monadic non-entailing operators and modifiers}, such as negation, modality, and non-subsective adjectives/adverbs.
Connectives have the type $<<e,t>, <<e,t>, <e,t>>>$, and we left open the choice between extending this to quantifiers and Pietroski's alternative account of quantification within conjunctivism.
Both predicate modifiers and connectives also require (re-)introducing function application to the system, albeit in a far more restricted manner than in the Montagovian framework.
Properly accounting for Content arguments would require much further discussion into the nature of propositions and intensionality; but here we simply adopted Pietroski's high-level account that some verbs take a Content argument instead of a Theme.

\section{The \eat format}
\label{sec:EAT}

In this section we provide a formal definition of \eat (Sections \ref{sec:EAT-definition}--\ref{sec:EAT-refinements}), and review its relation to prominent semantic representation formats used in NLP (Section \ref{sec:EAT-comparison}). For brevity, we use the term \emph{EAT} for both the representation format itself, and particular EAT-representations of sentences. When context does not clearly disambiguate the readings, we call the format the \emph{EAT-format}.

\subsection{Definition of EAT}
\label{sec:EAT-definition}

%

Starting with the transitive verb phrase (\ref{conj-UTAH}), the verb predicates an Event, and takes an Agent and Theme argument.
Based on this, we denote the semantic interpretations of the head, the external argument, and the internal argument as \emph{E}, \emph{A}, and \emph{T}, respectively.
All heads that project the syntactic structure (\ref{PS-schema}) are E-roles that take an A- and/or T-role as described in Section \ref{sec:conjunctivism}.
``E'', ``A'', and ``T'' are placeholders for more specific interpretations based on the lexical nature of the head (verb, preposition, connective).
\emph{Adjuncts} and \emph{functional heads} are monadic predicates that receive the default conjunctive interpretation (\ref{conj-conjunction}) (but see Sections \ref{sec:conjunctivism}; \ref{sec:EAT-refinements}).
They always modify a phrase that itself occupies the E-, A- or T-role in another configuration. Therefore, \emph{every element} in a sentence can be allocated to an E-, A-, or T-role, either directly or by virtue of modifying another element in that role.
We can now formulate a simple algorithm for allocating any syntactic element X to an E-, A-, or T-role based on its grammatical status, shown in (\ref{EAT-algorithm1}). E is a stand-alone role, while A and T are related to the E-role, marked by a subscript.

\begin{exe}
\ex \label{EAT-algorithm1}
\begin{itemize}
X and Y are syntactic elements.
\item[] if X is a verb, preposition, or connective:
\subitem{} ROLE(X) = E
\item[] if X = Int,Y:
\subitem{} ROLE(X) = T_{\text{Y}}
\item[] if X = Ext,Y:
\subitem{} ROLE(X) = A_{\text{Y}}
\item[] if X is a functional head or adjunct of (the extended projection of) Y:
\subitem{} ROLE(X) = ROLE(Y)
\end{itemize}
\end{exe}

%
%

We define an \emph{\eat-triplet} as a sequence of an E-, A-, and T-role in this order, such that the A- and T-roles are related to the E-role.
If a role is absent, we include a special empty token in its position, denoted as $\emptyset$.
Examples are shown in (\ref{EAT-ex3}).

\begin{exe}
\ex \label{EAT-ex3}
\begin{xlist}
\ex John sees Mary \\ $<$see, John, Mary$>$
\ex John walks \\ $<$walk, John, $\emptyset{}>$
\ex Mary was seen \\ $<$see, $\emptyset{}$, Mary $>$
\ex A cat on a roof \\ $<$on, cat, roof$>$
\end{xlist}
\end{exe}

Since adjuncts and functional heads modify an element that already occupies some role, we present these as further iterations of the same role in succession. Modifiers of each role can appear in the same \eat-triplet. If only some roles are modified, the rest receive the $\emptyset$-token.
Thematic arguments appear first, followed by the sequence of modifiers in the respective roles, as in (\ref{EAT-ex7}).

\begin{exe}
\ex \label{EAT-ex7}
Yesterday, a brown dog saw two white cats \\
$<$see, dog, cat$>$ $<$yesterday, brown, white$>$ $<\emptyset$, $\emptyset$, two$>$
\end{exe}

\emph{Possessors} are treated like other modifiers, but contain an additional possessive feature, as in (\ref{EAT-ex8}) (see Section \ref{sec:grammatical-features} for grammatical features). Possession could alternatively be represented with a preposition-esque metapredicate in the E-position, the possessor in the A-role and the possessed element in the T-role. However, we use the simpler semantically equivalent format to minimize the repetition of arguments in multiple positions.

\begin{exe}
\ex \label{EAT-ex8}
Mary's dog walks \\
$<$walk, dog, $\emptyset${}$>$ $<${}$\emptyset$, Mary$_{\text{POSS}}$, $\emptyset${}$>$
\end{exe}

The \eat of a whole sentence is made up by the sequence of all its \eat-triplets in a fixed order. Clausal arguments are represented by their main verb that appears first in the argument position and subsequently in its own E-role, as in (\ref{EAT-ex2}).\footnote{The current formalism does not explicitly mark the identity between words and their repetitions, as opposed to multiple instantiations of the same word in the original sentence. However, \emph{indexing} could be used for this, where each word token in the sentence would receive a unique index for marking token identity. While this is not a part of our current implementations (Sections \ref{sec:PCFG}--\ref{sec:Results}), its inclusion would be trivial by adding an index to the end of each word based on e.g. its position in the original sentence.}

\begin{exe}
\ex \label{EAT-ex2}
John sees that Mary greets Jim \\ $<$see, John, greet$>$ $<$greet, Mary, Jim$>$
\end{exe}

\eat thus has two distinct modes of combination: \emph{positional encoding} between the three roles within a single \eat-tuple, and \emph{concatenation} of multiple \eat-tuples.
These link to the types of combination recognized in the (basic) conjunctivist framework (Section \ref{sec:conjunctivism}).
Thematic arguments and dyadic relations (\ref{conj-UTAH})--(\ref{conj-PP}) are assigned by positional encoding, whereas the default conjunction interpretation (\ref{conj-conjunction}) corresponds to concatenating \eat-tuples.
Crucially, due to the extremely low number of semantic roles, \eat allows discarding metapredicates entirely, and instead using positional encoding for the (lemmatized) words themselves.
This allows trivial vectorization by replacing the words with embeddings (Section \ref{sec:EAT2seq}).

\subsection{Grammatical features}
\label{sec:grammatical-features}

In addition to thematic structure, we append the \eat-triplets with grammatical features.
We used the following features and values: \emph{force} (declarative/question/command), \emph{negation} (affirmed/negated), \emph{voice} (active/passive), \emph{tense} (present/past/perfect), \emph{aspect} (simple/progressive), \emph{number} (singular/plural), \emph{definiteness} (indefinite/definite), \emph{possessive} (non-possessive/possessive), and \emph{degree} (positive/comparative/superlative). Verbal features (force, negation, voice, tense, aspect) are specified for the E-role, nominal features (number, definiteness) for the A- and T-roles, and adjectival/adverbial features (degree) for all three roles.

Additionally, we allow a more generic representation of prepositions to minimize redundancy.
The standard representation discussed in \ref{sec:EAT-definition} above involves repetition, as the A-role of the preposition will also appear elsewhere in the sequence.
As another alternative, we replace the A-role with a marker of its prior role.
We add this marker as a grammatical feature of the preposition, and leave the A-role itself empty.
An example of both alternatives is shown in (\ref{EAT-ex5}).

\begin{exe}
\ex \label{EAT-ex5}
A dog in a house sees a cat on a roof \\
$<$see, dog, cat$>$ $<$in, dog, house$>$ $<$on, cat, roof$>$ \\
$<$see, dog, cat$>$ $<$in_{\text{A}} $\emptyset{}$, house$>$ $<$on_{\text{T}}, $\emptyset{}$, roof$>$
\end{exe}

We also use a similar technique for representing the referential links of \emph{relative pronouns} to the phrases they modify. A relative pronoun can relate to a prior A- or T-role, and we include this among nominal grammatical features. If a relative pronoun lacks both features, it has the default interpretation of modifying an entire prior \eat.
Like with prepositional arguments discussed above, an alternative would be to repeat the arguments themselves. The current implementation aims at minimizing redundancy, but both alternatives are available within syntactic parsing schemes that allow reconstructing the referential link. Examples are shown in (\ref{EAT-ex9}).

\begin{exe}
\ex \label{EAT-ex9}
\begin{xlist}
\ex John, who runs, sees Mary \\ $<$see, John, Mary$>$ $<$run, who_{\text{A}}, $\emptyset{}>$
\ex John sees Mary, who runs \\ $<$see, John, Mary$>$ $<$run, who_{\text{T}}, $\emptyset{}>$
\end{xlist}
\end{exe}

We do not use separate features for part-of-speech (POS), but it can be partly inferred from the grammatical features.
For example, all verbs are either active or passive, and hence if both voice features are $0$, the E-role is a \emph{non-verb}: a preposition(/connective), a modifier (e.g. an adverb), or the empty token.
Tense is represented by three binary features: present, past, and perfect. We refer to the present perfect as ``perfect'' and the past perfect as ``pluperfect''. If a verb lacks all tense features, it bears the infinitival inflection.
Our current implementation encodes the features with $28$ Boolean indicators ($0/1$) overall.
The \eat-format would easily allow features to be added or deleted based on task requirements (see Section \ref{sec:EAT-refinements}).

\subsection{Complex cases and possible refinements to \eat}
\label{sec:EAT-refinements}

In Section \ref{sec:conjunctivism-refinements} we reviewed three types of additions required to the basic conjunctivist system to account for a wider range of semantic phenomena: \emph{predicate modifiers}, \emph{connectives}, and \emph{Content arguments}.
We now review how these relate to \eat as defined in Sections \ref{sec:EAT-definition}--\ref{sec:grammatical-features}.
We argue that including markers for the relevant interpretations would be easy with minimal additions to \eat; but the information available from a surface-level syntactic parse alone is insufficient to warrant such decisions. Hence, further \emph{lexical} information would be needed, which we deliberately avoid relying on in order to allow the application of \eat to large datasets with dominant parsing schemes (Section \ref{sec:EAT-from-parse}).
We also make note of a similar issue with \emph{unaccusativity}, which \eat is already capable of representing but would require lexical specification to decide on.

\noindent{\textbf{Unaccusativity.}}
The subject of unaccusative verbs is the Theme even in the active voice \citep{Perlmutter78}:
e.g. \emph{The door opened} (cf. \emph{John opened the door}).
Unaccusatives could be similar to passives in lacking Ext,V in syntax, with Int,V raising to the grammatical subject position (e.g. \citealt{Hale:Keyser02, Pietroski05a}).
\eat already has the full capacity for representing verbs that lack the A-role, which in the active voice would result in the unaccusative reading.
Alternatively, unaccusatives could have a lexical origin prior to syntax (e.g. \citealt{Levin:Rappaport-Hovav1994}).
The most straight-forward implementation of this analysis in \eat would be to add another grammatical feature indicating the unaccusativity of the E-role.
However, whichever variant was chosen, a verb's unaccusativity would not be visible from surface-level syntax alone.

\noindent{\textbf{Predicate modifiers.}}
Currently, simple predicates and predicate modifiers are not distinguished in \eat: e.g. subsective and non-subsective adjectives are treated equally, resulting in false entailments with the conjunctive interpretation of \eat-sequences.
To remedy this, predicate modifiers ($<<e,t>, <e,t>>$) could be allocated separate Boolean features akin to the grammatical features in the \eat-tuple.
Each E-/A-/T-role would be connected to such a feature, and if its value was $1$, a monadic element (i.e. not a verb/preposition/connective) would be interpreted as a predicate modifier.
Like with unaccusativity, our reason for not implementing this was that such information is not available in the syntactic parse alone.
The problem could only be fixed by using lexical knowledge beyond the syntactic parse, which we deliberately abstained from.

\noindent{\textbf{Connectives.}}
If a separate feature was allocated for predicate modifiers as discussed above, one possible interpretation of this feature would be that it indicates \emph{second-order functions}.
Predicate modifiers are \emph{monadic} second-order functions from predicates to others; whereas connectives are \emph{dyadic} second-order functions from two predicates to a third.
(Here we rely on the analysis where clausal meanings are predicates instead of truth-values; see Section \ref{sec:conjunctivism-refinements}).
Hence, a single additional feature indicating a second-order reading could account for both predicate modifiers and connectives, relating to the latter in E-roles marked as prepositions in the current \eat.

\noindent{\textbf{Content vs. Theme role.}}
The current \eat treats clausal objects (of e.g. propositional attitude ascriptions) as normal T-roles that also head their own \eat-tuple as the E-role (Section \ref{sec:EAT-definition}). This results in a mistaken interpretation where the clausal object is entailed within the \eat it belongs to. In Section \ref{sec:conjunctivism-refinements} we simply adopted Pietroski's (\citeyear{Pietroski00, Pietroski05a}) high-level analysis of some verbs taking a Content argument instead of a Theme. As with unaccusativity and the second-order feature discussed above, we could add this simply as another Boolean feature to \eat.

\noindent{\textbf{Summary.}}
The \eat-format itself would be straight-forward to extend beyond simple conjunctive cases by adding Boolean indicators of second-order functions and Content roles.
Unaccusativity could be dealt with similarly, but can also be readily represented in the current system.
The issue with all these potential additions is not the capacity of \eat, but instead the availability of information beyond the kinds of syntactic parses we assume to be available as input to \eat-construction (Section \ref{sec:EAT-from-parse}).
We therefore did not include them to the present implementations of \eat.
Crucially, a comparable issue arises with \emph{any} semantic parsing framework limited in such a way.
Our discussion here provided instructions for incorporating further information to \eat in possible future work.
In comparison to alternative semantic frameworks (Section \ref{sec:EAT-comparison}), such changes would be maximally simple to include in \eat, adding only a single bit of information ($0/1$) per feature.

\subsection{Comparison to alternative semantic representation formats}
\label{sec:EAT-comparison}

While \eat is unique in being directly inspired by conjunctivism (Section \ref{sec:conjunctivism}), Neo-Davidsonian event semantics has been applied in prior logical form implementations for NLP \citep{Bos2015, Reddyetal2016, Reddyetal2017}.
There are also interesting affinities between \eat and \emph{Discourse Representation Theory} (DRT) \citep{Kamp81, Kamp95}. For example,
existential quantification and conjunction have important roles as default interpretations in both.
Neo-Davidsonian variants of DRT also exist \citep{Bos2015}.
If this comparison was taken further, it might be feasible to interpret \eat as a simplified representation of (some version of) DRT.
However, a systematic analysis of conjunctivism in relation to DRT has so far been lacking, and is beyond our present scope.

%
%
Another prominent formalism is \emph{Abstract Meaning Representation} (AMR) \citep{Banarescuetal13},
which is a graph-based representation of variables (nodes), concepts that predicate them (node labels) and semantic relations between variable (edges). The relations it recognizes include argument slots from PropBank \citep{Palmeretal05}, additional semantic roles (e.g. \emph{beneficiary} or \emph{destination}), negation, modality, etc. AMR recognizes $\sim 100$ semantic relations overall, and abstracts away from most grammatical information (e.g. tense).
It also assimilates between some syntactically divergent expressions with logically equivalent interpretations. Using an example from \citet{Banarescuetal13}, the following sentences all have the same AMR analysis: (i) \emph{he described her as a genius}, (ii) \emph{his description of her: genius}, and (iii) \emph{she was a genius, according to his description}.
Both MRS (see below) and \eat remain closer to the surface syntax than AMR.

\emph{Minimal Recursion Semantics} (MRS) maintains more proximity to the surface grammar than AMR \citep{Copestakeetal2005, Copestake2009}.
First, MRS deliberately retains ambiguity when disambiguation is not possible from the syntactic parse alone.
Second, it contains more grammatical information than AMR (e.g. tense).
Finally, semantic roles in MRS are not specified for pre-existing lexical information like PropBank frames. Hence, the roles have less semantic detail than in AMR; but this simplifies the link between the syntactic parse and MRS.
Empirically, MRS has demonstrated superior results to AMR in parsing and text generation using encoder-decoder techniques.
\citet{Lin:Xue2019} account for this on three main grounds: (i) AMR's higher degree of abstraction from surface forms, (ii) AMR's finer-grained classification of named entities, and (iii) MRS's semantic roles bearing a closer relation to syntactic roles than AMR's.

MRS can be seen as an attempt to bring semantic parsing closer to the syntactic parse for optimizing the simplicity of mapping between them.
However, like AMR, it continues to rely on a large number of semantic roles. For example, the dataset used in Hajdik et al's (\citeyear{Hajdiketal2019}) experiments has $52$ roles (Section \ref{sec:comparison-mrs}).
This is a stark contrast to \eat, the main motive of which is to reduce the roles to the bare minimum.
In Section \ref{sec:comparison-mrs} we demonstrate that drastically reducing the roles improves text reconstruction from MRS.
Moreover, the $50-100$ roles of AMR and MRS are obviously not tailored for \emph{positional encoding} like the three roles of \eat.
Hence, \eat is unique in discarding metapredicates for semantic roles and instead allocating them dedicated positions.
This makes \eat much easier to manage for a variety of NLP tasks (see Section \ref{sec:Results}).
Overall, compared to AMR and MRS, we consider the main benefits of \eat to be its \emph{simplicity} and \emph{versatility}.

\section{Obtaining \eat from a syntactic parse}
\label{sec:EAT-from-parse}

In this section we present a mapping from syntactic structure to \eat.
We apply this for both \emph{probabilistic context-free grammar} (PCFG) (Section \ref{sec:PCFG}) and \emph{dependency grammar} (Section \ref{sec:dependency-EAT}), using no external information aside of the (morpho)syntax.
In notation, we mark the E-role that an A- or T-role relates to with a subscript when there is possible ambiguity.


\subsection{Mapping rules between PCFG and \eat}
\label{sec:PCFG}

We go through the rules we used for constructing an \eat from PCFG parse as implemented in the Stanford PCFG parser \citep{Klein:Manning03}.\footnote{\url{https://nlp.stanford.edu/software/lex-parser.shtml}} In addition to phrase structural information, it also includes POS-tags in the Penn Treebank notation  \citep{Taylor2003}.

\subsubsection{Phrase heads}
\label{sec:phrase-structure-heads}

We map each non-terminal phrase to a terminal word. To achieve this we first map each phrase to its daughter that specifies its syntactic nature. These can be nonterminal. Here, we use ``head'' to refer to all such elements, and ``terminal head'' for word-level heads.
Applying the labeling recursively, we ultimately map each non-terminal phrase to a terminal head.

First, we link each POS-tag to the word it tags. We then map phrases to POS-tags that they dominate. This resembles the relation between an X'-level phrase and its head in X'-theory (Section \ref{sec:argument-structure-syntax}). The possible tags determining the head of each phrase type are specified in (\ref{PS-heads}).

\begin{exe}
\ex \label{PS-heads}
\begin{itemize}
\item[] S/SQ/SINV $\rightarrow$ $\{$VP$\}$
\item[] SBAR/SBARQ $\rightarrow$ $\{$S, SQ$\}$
\item[] NP $\rightarrow$ $\{$NN, NNP, NNPS, NNS$\}$
\item[] VP $\rightarrow$ $\{$VB, VBD, VBG, VBN, VBP, VBZ$\}$
\item[] ADJP $\rightarrow$ $\{$JJ, JJR, JJS$\}$
\item[] ADVP $\rightarrow$ $\{$RB, RBR, RBS$\}$
\item[] PP/WHPP $\rightarrow$ $\{$IN, TO$\}$
\item[] CONJP $\rightarrow$ $\{$CC, IN$\}$
\item[] WHNP $\rightarrow$ $\{$WP, WDT, NN, NNP, NNPS, NNS$\}$
\item[] WHADJP/WHADVP $\rightarrow$ $\{$WRB, JJ, RB$\}$
\end{itemize}
\end{exe}

Sentences (S/SINV/SQ) are exceptional in that they are not directly labeled by POS-tags but by a \emph{verb phrase}. This implements the common theoretical notion that a sentence is an extended projection of its main verb \citep{Grimshaw91, Grimshaw05}.
%
If a phrase does not dominate any of the heads in (\ref{PS-heads}) but dominates another instance of the same phrase type, we treat the lower phrase as the head. For VPs we exceptionally apply this rule before searching for a lexical head in (\ref{PS-heads}), as a VP dominating another VP designates an auxiliary-VP construction in PCFG. Instead of treating the auxiliary as the head (as applying (\ref{PS-heads}) initially would), we find the head verb in the lower VP, leaving the auxiliary as a modifier (see Section \ref{sec:phrase-structure-argument} for arguments and modifiers).

With other phrases types, the recursive embedding rule is applied only if (\ref{PS-heads}) fails to find a head.
If a phrase dominates multiple heads of the same type, this instantiates either \emph{(head) conjunction} or \emph{compounding}. If the heads are separated by a conjunction, we treat the first as the phrase head (e.g. \emph{\textbf{dogs} and cats}). If not, the phrase is a compound and we treat the last as the phrase head (e.g. \emph{dog \textbf{lover}}).

Once we have mapped each non-terminal phrase to a head, we then apply the mapping recursively as long as the head of a phrase is mapped to another head, until each non-terminal phrase is mapped to a terminal head. Hence, sentences and VPs get mapped to verbs, NPs to nouns etc. This allows us to eliminate non-terminal phrases from \eat.

\subsubsection{Argument structure}
\label{sec:phrase-structure-argument}

We first find arguments for non-terminal phrases, which we then map to terminal heads as specified in Section \ref{sec:phrase-structure-heads}. For example, the subject is first assigned for sentences (S/SINV/SQ), and the object for VPs. However, both the sentence and the VP are mapped to the same terminal head: the main verb. Therefore, the main verb will ultimately include both the subject and object arguments. The main verb of a sentence always occupies the E-role of the first triplet in the \eat-sequence (Section \ref{sec:EAT}).


\noindent{\textbf{Clauses.}}
There are four types of clauses in PCFG: \emph{S}, \emph{SQ}, \emph{SINV}, \emph{SBAR}, and \emph{SBARQ}. Of these, S is the basic declarative clause, SQ is a question, and SINV is a declarative with auxiliary inversion.
When S(/SINV) contains a VP, we treat this VP as the head.
Recursively, the V-head of the VP will thus become the head of the whole sentence.
If a NP precedes the head VP, we map it to the A-role if the verb has active voice and the T-role if the verb has passive voice, as shown in (\ref{PS-S}) (see Section \ref{sec:phrase-structure-grammar} for voice determination). Comparing to the syntactic analysis of Section \ref{sec:argument-structure-syntax}, these PCFG-constructions roughly correspond to the TP with the grammatical subject in Spec,TP; raised there from the original thematic position in the VP.

\begin{exe}
\ex \label{PS-S}
\begin{minipage}[t]{0.4\textwidth}
\Tree [.S NP\\(A) [.VP [.(...) V\\{[active]} ]]]
\end{minipage}
\begin{minipage}[t]{0.4\textwidth}
\Tree [.S NP\\(T) [.VP [.(...) V\\{[passive]} ]]]
\end{minipage}
\end{exe}

We first map all NPs preceding an active VP to the A-role. Then, if two elements were initially mapped to the A-role, the first of which is a \emph{wh}-phrase (e.g. \emph{who}, \emph{what}), we change this \emph{wh}-phrase to the T-role. This is an English-specific aspect of word order. When the subject is a \emph{wh}-phrase, the active verb precedes the object as in declaratives. Hence, if \emph{both} the subject and a \emph{wh}-phrase precede the verb in an active voice question, the \emph{wh}-phrase must be an object.
Simplified PCFG-parses of two examples are shown in (\ref{PS-SQ}). In the analysis of Section \ref{sec:argument-structure-syntax}, the \emph{wh}-phrase would  occupy Spec,CP above the TP.

\begin{exe}
\ex \label{PS-SQ}
\begin{minipage}[t]{0.4\textwidth}
\Tree [.SQ Who\\(A) [.VP \edge[roof]; {saw Mary} ]]
\end{minipage}
\begin{minipage}[t]{0.4\textwidth}
\Tree [.SQ Who\\(T) did John\\(A) [.VP \edge[roof]; {see} ]]
\end{minipage}
\end{exe}

SBAR and SBARQ represent embedded clauses.
In principle, SBAR-sentences are declarative and SBARQ-sentences questions, but the SNLI dataset contains some exceptions to this. Our mapping rules are sufficiently robust not to be vulnerable to such variation in PCFG. SBAR branches to a relative pronoun or a complementizer on the left and a sentence on the right, roughly corresponding to CPs with a relative pronoun at Spec,CP or complementizer at C (Section \ref{sec:argument-structure-syntax}).
We treat relative pronouns in the same way as \emph{wh}-phrases in questions (see above). The corresponding ordering constraint appears here: if both a relative pronoun and another phrase are initially mapped to the A-role, the relative pronoun is reassigned to the T-role. 

Gerundive VPs can take Agent NP arguments. These are sometimes treated as regular clauses where the V-head is inflected with the gerundive tag (VBG). However, some gerundive constructions are instead parsed as nested NPs with a VP modifier headed by VBG. We assimilate the latter to the corresponding clauses, with the NP head serving as the A-role of the gerundive verb. We show these alternative variants of the same gerundive construction in (\ref{PS-GER}a--b).

\begin{exe}
\ex \label{PS-GER}
\begin{minipage}[t]{0.4\textwidth}
\Tree [.S NP\\{a man}\\(A) [.VP VBG\\drinking\\(E) NP\\tea\\(T) ]]
\end{minipage}
\begin{minipage}[t]{0.4\textwidth}
\Tree [.NP NP\\{a man}\\(A) [.VP VBG\\drinking\\(E) NP\\tea\\(T) ]]
\end{minipage}
\end{exe}

\noindent{\textbf{Verb phrases.}}
As an object argument initially mapped to the T-role, we allow the following VP-daughters: noun phrases (NP), sentences (S, SQ, SBAR, SBARQ), and adjective phrases (ADJP).
These correspond to Int,V in the framework of Section \ref{sec:argument-structure-syntax}.
If a verb has two arguments initially mapped to the T-role, we then change the first of these (in linear word order) to the Recipient argument, which we denote by the preposition \emph{to} (Section \ref{sec:conjunctivism-phrases}). Examples are shown in (\ref{PS-VP1}).

\begin{exe}
\ex \label{PS-VP1}
\begin{minipage}[t]{0.4\textwidth}
\Tree [.VP V\\saw\\(E) NP\\Mary\\(T) ]
\end{minipage}
\begin{minipage}[t]{0.4\textwidth}
\Tree [.VP V\\gave\\(E) NP\\Mary\\(T_{\text{to}}) NP\\flowers\\(T_{\text{gave}}) ]
\end{minipage}
\end{exe}

%
%

Infinitival clauses belong to the S-category, and contain a VP introduced by the infinitival \emph{to} (with the POS-tag TO). If an infinitival clause is the T-role of a verb and lacks an A-role, it inherits the A-role from the verb, as in (\ref{PS-VP3}).

\begin{exe}
\ex \label{PS-VP3}
\Tree [.S John\\(A_{\text{wants}})\\(A_{\text{run}}) [.VP V\\wants\\(E) [.S [.VP TO\\to [.VP V\\run\\(T_{\text{wants}})\\(E) ]]]]]
\end{exe}

\noindent{\textbf{Preposition/complementizer phrases.}}
Compared to the discussion in Section \ref{sec:argument-structure-syntax}, PCFG is close to the adjunct analysis of PPs, where they are modifiers of the external argument, as in (\ref{PS-PP}).

\begin{exe}
\ex \label{PS-PP}
\begin{minipage}[t]{0.4\textwidth}
\Tree [.NP N\\dog\\(A_{\text{in}}) [.PP P\\in\\(E) [.NP \edge[roof]; {a house\\(T_{\text{in}})} ]]]
\end{minipage}
\begin{minipage}[t]{0.4\textwidth}
\Tree [.VP V\\live\\(A_{\text{in}}) [.PP P\\in\\(E) [.NP \edge[roof]; {a house\\(T_{\text{in}})} ]]]
\end{minipage}
\end{exe}

In passive verbs and adjectival participles (both tagged VBN), the Agent argument is marked with a PP headed by the preposition \emph{by}. We therefore change \emph{by}-PPs to the A-role if the verb's POS-tag is VBN and the P-complement is a NP.

We also assimilate complementizers like \emph{while} to prepositions that take a clause instead of an NP as the T-role.
This fits well with PCFG's POS-tagging scheme that assigns both prepositions and complementizers to the same tag (IN).
The application of our complementizer rule is demonstrated in (\ref{PS-CP}).

\begin{exe}
\ex \label{PS-CP}
\Tree [.S NP\\John\\(A_{\text{talks}}) [.VP V\\talks\\(E)\\(A_{\text{while}}) [.SBAR C\\while\\(E) [.S NP\\Mary\\(A_{\text{walks}}) [.VP V\\walks\\(T_{\text{while}})\\(E) ]]]]]
\end{exe}

We further treat \emph{conjunctions} (CC) as analogical to P/C. This is in line with syntactic proposals where conjunctions have been brought under the X'-schema via conjunction phrases (ConjP), conjuncts occupying Spec,ConjP and Comp,ConjP \citep{Kayne94, Johanessen98}. PCFG follows a more traditional analysis of conjunction constructions as ternary branching trees, as in (\ref{PS-Conj}).

\begin{exe}
\ex \label{PS-Conj}
\Tree [.NP [.S \edge[roof]; {John talks}\\(A_{\text{and}}) ] CC\\and [.S \edge[roof]; {Mary walks}\\(T_{\text{and}}) ]]
\end{exe}


\noindent{\textbf{Modifiers.}}
Remaining words/phrases immediately dominated by a phrase are mapped to the default modifier role of the (head of the) phrase.
We concatenate the modifiers of each E/A/T-role to the corresponding role position  (Section \ref{sec:EAT}).

\subsection{Mapping rules between dependency grammar and \eat}
\label{sec:dependency-EAT}

In this section we go through the mapping rules between dependency parses and \eat.
A \textit{dependency graph} contains information about three kinds of properties of a word: (i) its intrinsic features (e.g. POS-tag, lemma, and inflection), (ii) its \textit{head} (of which it is a \textit{dependent}), and (iii) the nature of the dependency relation it bears to its head.

%

\noindent{\textbf{Clauses.}}
Active sentences can have a subject (\textit{nsubj}) and an object (\textit{dobj}). We map \textit{nsubj} to the A-role and \textit{dobj} to the T-role, as in (\ref{thematic1}).

\begin{exe}
\ex \label{thematic1}
\begin{dependency}
\begin{deptext}[column sep=0.4cm]
Mary \& makes \& a \& sandwich \\
A \& E \& \& T \\
\end{deptext}
\depedge{1}{2}{nsubj}
\depedge{4}{2}{dobj}
\end{dependency}
\end{exe}

Passive clause subjects have a separate relation to the verb (\textit{nsubjpass}). If a direct object (\textit{dobj}) is present, the passive subject is a Recipient, which we treat as the T-role of an implicit preposition \textit{to} modifying the verb (Section \ref{sec:conjunctivism-phrases}), shown in (\ref{thematic2}).

\begin{exe}
\ex \label{thematic2}
\begin{dependency}
\begin{deptext}[column sep=0.4cm]
John \& is \& sung \& a \& song \\
T_{\text{to}} \& \& E \&\& T_{\text{sung}} \\
\end{deptext}
\depedge{1}{3}{nsubjpass}
\depedge{5}{3}{dobj}
\end{dependency}
\end{exe}

Otherwise, passive subjects are assigned to the T-role of the verb. Agents introduced in passive constructions via \textit{by} are dependents of this preposition, which relates to the verb via the \textit{agent} relation in the dependency graph. We assign them to the A-role, as in (\ref{thematic3}).

\begin{exe}
\ex \label{thematic3}
\begin{dependency}
\begin{deptext}[column sep=0.4cm]
a \& sandwich \& is \& eaten \& by \& Mary \\
\& T \& \& E \& \& A \\
\end{deptext}
\depedge{2}{4}{nsubjpass}
\depedge{6}{5}{pobj}
\depedge{5}{4}{agent}
\end{dependency}
\end{exe}

In addition to standard nominal arguments, there are clausal subjects (\textit{csubj}, \textit{csubjpass}), clausal objects (\textit{ccomp}, \textit{xcomp}, \textit{attr}, \textit{oprd}), and adjectival (predicative) objects (\textit{acomp}). We assign these to A- and T-roles similarly to nominal arguments.
Like with PCFG before (Section \ref{sec:PCFG}), clausal arguments project their own A- and T-roles, resulting in their repetition both in the embedded argument positions and as subsequent E-roles.
An embedded clause example is provided in (\ref{thematic4})

\begin{exe}
\ex \label{thematic4}
\begin{dependency}
\begin{deptext}[column sep=0.5cm]
John \& saw \& that \& Mary \& became \& tired \\
A_{\text{saw}} \& E \& \& \& T_{\text{saw}} \\
\&\&\& A_{\text{became}} \& E \& T_{\text{became}} \\
\end{deptext}
\depedge{1}{2}{nsubj}
\depedge{4}{5}{nsubj}
\depedge{5}{2}{ccomp}
\depedge{6}{5}{acomp}
\end{dependency}
\end{exe}

Unlike PCFG parses, dependency graphs allow syntactic relations to be preserved across clause types with word order alterations. For instance, \textit{wh}-phrases raised to the clause-initial position in questions or relative clauses retain the dependency relation that determines their argument status with respect to the verb.
Hence, \emph{what} in (\ref{thematic5}) remains a \emph{dobj}.

\begin{exe}
\ex \label{thematic5}
\begin{dependency}
\begin{deptext}[column sep=0.3cm]
what \& did \& John \& see? \\
T \& \& A \& E \\
\end{deptext}
\depedge{1}{4}{dobj}
\depedge{3}{4}{nsubj}
\end{dependency}
\end{exe}

\noindent{\textbf{Preposition/complementizer phrases.}}
Prepositions are marked with the \textit{prep} role and are dependents of the element the prepositional phrase modifies, assigned to the A-role. The complement of the preposition is marked as \textit{pcomp}, and we allocate it to the T-role.
Some prepositional phrases are ambiguous between multiple interpretations, such as \textit{in the street} in (\ref{thematic6}), which could modify either the main verb (\textit{greets}) or the object noun (\textit{woman}). However, such ambiguities are already resolved at the syntactic parsing stage (the verb being chosen here).
As before, we also assimilate \emph{complementizers} and \emph{connectives} to prepositions (Section \ref{sec:conjunctivism}).

\begin{exe}
\ex \label{thematic6}
\begin{dependency}
\begin{deptext}[column sep=0.3cm]
A \& man \& with \& a \& bag \& greets \& a \& woman \& in \& the \& street \\
\& A_{\text{greets}} \&\&\&\& E \& \& T_{\text{greets}} \\
\& A_{\text{with}} \& E \& \& T_{\text{with}} \& A_{\text{in}} \&\&\& E \&\& T_{\text{in}} \\
\end{deptext}
\depedge{2}{6}{nsubj}
\depedge{3}{2}{prep}
\depedge{5}{3}{pobj}
\depedge{8}{6}{dobj}
\depedge{9}{6}{prep}
\depedge{11}{9}{pobj}
\end{dependency}
\end{exe}

%

\noindent{\textbf{Modifiers.}}
We assign non-prepositional dependents to the default modifier role, all modifiers being concatenated to their corresponding role position in the resulting \eat-sequence (Section \ref{sec:EAT}).
Since we applied the \eat derived from the dependency graph to text reconstruction and transformation (Section \ref{sec:EAT2seq}), we further concatenated the modifiers in a canonical order to ensure their independence of surface word order, which may vary across clause types. We recognize the following modifier types in the corresponding order: \emph{determiners} (except articles), \emph{modal auxiliaries}, \emph{adjectives}, \emph{adverbs}, \emph{negations} (except in verbs), \emph{numerals}, \emph{appositives}, \emph{compounds}, and \emph{possessors}. We detect these using the dependency relation and POS-tag features in the dependency graph. Dependents that are not in any of these classes are assigned to the default class of remaining modifiers, which comes last in the concatenation order.

\subsection{Grammatical features}
\label{sec:phrase-structure-grammar}

We mapped each word to a list of grammatical features based on its POS-tag and syntactic context. Some features can be found directly in the POS-tag, but others require reference to the surrounding context. When the semantic interpretation of a grammatical element (e.g. negation, auxiliary, or article) was fully included in a grammatical feature, we removed the element from the modifier list in \eat to eliminate redundancy.
Below, we present our rules for obtaining grammatical features in a generic form applicable to both PCFG and dependency grammar.

\noindent{\textbf{Force.}}
We treated a clause as a \emph{question} if its grammatical subject (A-role in actives and T-role in passives) was preceded by an auxiliary or the copula verb (\textit{be}).\footnote{PCFG uses the tag \emph{SQ} for questions, but on manual evaluation this was unreliable in the SNLI dataset (Section \ref{sec:parallel-corpora}).} If a verb lacked both a grammatical subject and tense inflection, we treated it as a \emph{command}.

\noindent{\textbf{Negation.}}
Negated verbs have \emph{not/n't} as a modifier.

\noindent{\textbf{Tense.}}
We stored features for \emph{present}, \emph{past}, and \emph{perfect} tense.
We assigned tense based on the POS tag of the verb or its auxiliary: VBG/VGP/VBZ for present and VBD/VBN for past.
A \emph{have}-auxiliary together with the verb's participle tag (VBN) results in the perfect tense, and the \emph{have}-auxiliary's tense distinguishes between present perfect (called simply \emph{perfect}) and \emph{pluperfect}.

\noindent{\textbf{Aspect.}}
\emph{Progressive} verbs have the gerund tag (VBG) and a copula auxiliary (\emph{be}).

\noindent{\textbf{Voice.}}
\emph{Passive} verbs are inflected with the participle tag (VBN) and modified by a copula auxiliary (\emph{be}) that determines the tense.

\noindent{\textbf{Definiteness.}}
\emph{Definite} nouns have \textit{the} as a modifier.

\noindent{\textbf{Number.}}
\emph{Plural} nouns have NNS or NNPS as the POS-tag.

\noindent{\textbf{Possessive.}}
Possessors are modified by the \emph{'s} clitic in PCFG, and have the dependency relation \emph{poss} in dependency grammar. We also assigned the irregular possessive inflections of personal pronouns (\emph{my}, \emph{your}, etc.) to the possessive feature via separate manually programmed rules.

\noindent{\textbf{Degree.}}
\emph{Comparative} adjectives are tagged as JJR, \emph{superlative} adjectives as JJS, comparative adverbs as RBR, and superlative adverbs as RBS.
Another way of marking comparison class is via the modifier \emph{more/most}, which we assimilated to the degree feature in \eat.
\section{Experiments}
\label{sec:Results}

We now move on to applications of \eat.
We first show that text generation performance from MRS can be improved by significantly reducing the number of semantic roles, inspired by \eat (Section \ref{sec:comparison-mrs}).
Even a drastic reduction of the number of roles to the absolute minimum of \emph{three} can still largely maintain performance.
Then, we use \eat to obtain \emph{parallel corpora} between grammatical classes from both PCFG- and dependency-parsed datasets (Section \ref{sec:parallel-corpora}).
Finally, we generate English text from \eat-input, and apply this to \emph{grammatical transformation} in $14$ directions (Section \ref{sec:EAT2seq}).
We make all source code and generated corpora available as open-source.

%

\subsection{Reducing semantic roles in MRS for text reconstruction} 
\label{sec:comparison-mrs}

In this section we empirically demonstrate that limiting the range of arguments in semantic representations along the lines specified in Sections \ref{sec:Background}--\ref{sec:EAT} can benefit \emph{text reconstruction}. As our starting point, we used the linearized implementation of MRS (see Section \ref{sec:EAT-comparison}) from \citet{Hajdiketal2019}. We then modified MRS to use significantly fewer argument classes, assimilating semantic roles as motivated by our theoretical framework (Section \ref{sec:Background}). We retained everything else in the original MRS to ensure that the differences arose only from these alterations of roles. Hence, here we did not yet implement the \eat-format as such (Section \ref{sec:EAT}), but rather a more \eat-esque variant of MRS that we call \emph{MRS-EAT}.

\subsubsection{Data}
We used the scripts from Hajdik et al.'s (\citeyear{Hajdiketal2019}) GitHub repository,\footnote{\url{https://github.com/shlurbee/dmrs-text-generation-naacl2019}} which also loads the training, validation, and test data derived from the Redwoods Treebank corpus \citep{Oepenetal2002}.\footnote{\url{http://svn.delph-in.net/erg/tags/1214/tsdb/gold}} The training set has $67313$ sentences, the validation set $5288$, and the test set $10201$.\footnote{That loaded training set was smaller than the training set reported in \citet{Hajdiketal2019}, which had 72190 sentences. This is likely due to the filtering of problematic parses (e.g. parenthesis mismatches) or the exclusion of overlapping elements with the validation or test sets, both automatically conducted at the data collection and pre-processing stages of the script.}

\subsubsection{Approach}

\emph{Dependency MRS} (DMRS) is a graph-representation of MRS, with nodes designating word meanings and edges their semantic relations \citep{Copestake2009}.
\citet{Hajdiketal2019} used DMRS linearized in the Penman format \citep{Goodman2020} for text generation via \emph{neural machine translation} (NMT), training the system to reconstruct original English sentences.
Compared to AMR-based reconstruction in the same test settings \citep{Konstas2017}, they achieved significantly better performance ($75.8$ vs. $33.8$ BLEU-score).

We replicated the experiments of \citet{Hajdiketal2019}. The DMRS-representations contained $52$ possible semantic relations altogether, which is why we denote it as \emph{MRS_{52}}. To shift the representation closer to \eat without changing the Penman format itself, we kept everything intact except for the role labels, which we reduced based on the theoretical discussion in Sections \ref{sec:Background}--\ref{sec:EAT}.
We call MRS with \eat-inspired reduced arguments \emph{MRS-EAT}.

MRS is \emph{hierarchical}, and all arguments (including modifiers) are dominated by their head.
The E-role is therefore implicit, as anything dominating an A- or T-argument is automatically its E-role.
On the other hand, this also requires a separate \emph{M-role} for \emph{modifiers}.
We treated every argument that was not the A- or T-role as a modifier, except \emph{prepositions} which we allocated to a distinct role.
We also separated \emph{clausal} A- and T-roles from other variants, and the T-role of a preposition from the rest.
Finally, we distinguished the left- and right-arguments of \emph{connectives} (L-CONN and R-CONN) from A- and T-roles.
This variant of MRS-\eat thus had \emph{nine} roles altogether, and we call it \emph{MRS-\eat{}_9}.
We additionally experimented on a maximally reduced variant of MRS-\eat that only had \emph{three} roles: A, T and M. We call this variant \emph{MRS-\eat{}_3}.
Table \ref{tab:MRS-EAT-conversion} demonstrates the conversion from MRS_{52} to both variants of MRS-\eat.

\begin{table}[t]
\begin{center}
\begin{tabular}{|l|c|c|} \hline
\textbf{MRS_{52}} & \textbf{MRS-\eat{}_9} & \textbf{MRS-\eat{}_3} \\ \hline
ARG1-NEQ & A & \multirow{5}{*}{A} \\ \cline{1-2}
ARG1-H & A-CLAUSE & \\ \cline{1-2}
L-INDEX-NEQ & \multirow{3}{*}{L-CONN} & \\ \cline{1-1}
L-HNDL-HEQ & & \\ \cline{1-1}
L-HNDL-H & & \\ \hline
ARG2-NEQ (of verbs) & T & \multirow{6}{*}{T} \\ \cline{1-2}
ARG2-NEQ (of prepositions) & T-PREP & \\ \cline{1-2}
ARG2-H & T-CLAUSE & \\ \cline{1-2}
R-INDEX-NEQ & \multirow{3}{*}{R-CONN} & \\ \cline{1-1}
R-HNDL-HEQ & & \\ \cline{1-1}
R-HNDL-H & & \\ \hline
prepositions (POS-tag ``p'') & P & \multirow{2}{*}{M} \\ \cline{1-2}
rest & M & \\ \hline
\end{tabular}
\caption{Conversion from MRS\textsubscript{52} to MRS-\eat{}\textsubscript{9} and MRS-\eat{}\textsubscript{3}.}
\label{tab:MRS-EAT-conversion}
\end{center}
\end{table}

For training the model, the script uses OpenNMT \citep{Kleinetal2017}.\footnote{\url{https://opennmt.net/}} We replicated the best-performing model architecture from \citet{Hajdiketal2019}. The initial embedding layer had 500 dimensions, and both the encoder and decoder were two-layered 800-dimensional \emph{LSTM} networks \citep{Hochreiter:Schmidhuber97}. Training used the negative log-likelihood loss function, a dropout probability of $0.3$, and the Adam optimizer \citep{ADAM14} initialized with a $0.001$ learning rate.
We trained the model with each of the three formats: MRS_{52}, MRS-\eat{}_9, and MRS-\eat{}_3. With a batch size of $50$, each model's validation accuracy stabilized or decreased after $14$ training epochs. The number of training iterations was thus identical in every case.

\subsubsection{Results}
\label{sec:MRS-results}

Table \ref{tab:MRS} shows the performance of the NMT models trained on MRS_{52}, MRS-\eat{}_{9}, and MRS-\eat{}_{3} on five metrics.
\emph{BLEU} \citep{Papinenietal02a} is a common means of evaluating machine translation based on n-gram overlap between the candidate translation and a gold standard.
We distinguish between n-grams of length $1-4$.
\emph{METEOR} \citep{Banerjee:Lavie2005} is also based on n-gram overlap, but additionally considers paraphrases and synonyms from WordNet \citep{Wordnet95}.
For both BLEU and METEOR, we used implementations from \textit{nlg-eval}.\footnote{\url{https://github.com/Maluuba/nlg-eval}}
Finally, we also measured the ratio of \emph{exact matches}, i.e. generated texts that were identical with the original. We lowercased both original and generated texts prior to all evaluation.

All three techniques reached similar scores, differences never exceeding four points in BLEU or METEOR.
More surprisingly, MRS-\eat_9 \emph{increased} the performance from MRS_{52} in all metrics: BLEU by $0.38-0.54$ points, METEOR by $0.45$, and exact matches by $3.6\%$.
These results illustrate that appropriate reduction of semantic roles can not only significantly simplify the representation, but \emph{improve} text reconstruction.
A likely reason for this is the increased \emph{generalizability} of the model, which can reduce overfitting to the training data.
Also, even though MRS-\eat{}_3 had lower performance than MRS_{52}, the reduction remained minor: BLEU diminished only by $1.88-3.16$ points, METEOR by $1.64$, and exact matches by $2.3\%$.

\begin{table}[t]
\begin{center}
\begin{tabular}{|c|c|c|c|c|c|c|}
\hline
\textbf{Technique} & \textbf{BLEU-1} & \textbf{BLEU-2} & \textbf{BLEU-3} & \textbf{BLEU-4} & \textbf{METEOR} & \textbf{Exact match} \\ \hline
MRS_{52} & $83.28$ & $75.82$ & $69.66$ & $64.22$ & $47.60$ & $20.2\%$ \\
MRS-\eat{}_9 & $\mathbf{83.66}$ & $\mathbf{76.23}$ & $\mathbf{70.13}$ & $\mathbf{64.76}$ & $\mathbf{48.05}$ & $\mathbf{23.8\%}$ \\
MRS-\eat{}_3 & $81.40$ & $73.30$ & $66.75$ & $61.06$ & $45.96$ & $17.9\%$ \\ \hline
\end{tabular}
\caption{Comparison between MRS\textsubscript{52}, MRS-\eat{}\textsubscript{9}, and MRS-\eat{}\textsubscript{3} in text generation performance.}
\label{tab:MRS}
\end{center}
\end{table}

\subsubsection{Summary}
We conclude that optimizing the number and selection of semantic roles deserves more focus in future work on semantic representation.
When reducing the number of roles from prior formats (like MRS_{52}), it is crucial that they are appropriately chosen to optimize expressiveness.
The success of MRS-\eat{}_{9} indicates that its selection of roles is beneficial -- possibly even \emph{optimal} -- for text generation.
Furthermore, even though MRS-\eat{}_{3} had lower performance than MRS_{52}, the differences were relatively minor in comparison to the scale of the role reduction.
Clearly the vast majority of information was still retained even when only \emph{three} roles were used as opposed to $52$.
We take this to motivate investigating how far it is possible to go with only three roles plus some additional grammatical information.
This is the main goal of \eat.

\subsection{Constructing parallel corpora between grammatical classes}
\label{sec:parallel-corpora}

In this section we demonstrate how \eat can be used to build parallel corpora between grammatical classes from a syntactically parsed monolingual corpus.
Parallel sentences are identical in argument structure, and only differ in grammar.

\subsubsection{Data}
We generated the parallel corpora from two datasets parsed via PCFG and dependency grammar, respectively.
First, we used the \emph{Stanford Natural Language Inference} (SNLI) corpus \citep{Bowmanetal2015},\footnote{\url{https://nlp.stanford.edu/projects/snli/}} which is pre-parsed with the Stanford PCFG parser \citep{Klein:Manning03}.
Second, we used English sentences derived from \textit{Stanford NMT parallel corpora} (SNMT) \citep{Luongetal2015, Luong:Manning2016}.\footnote{\url{https://nlp.stanford.edu/projects/nmt/}} which we parsed ourselves using SpaCy.\footnote{\url{https://spacy.io/}} We limited SNMT to sentences that had \mbox{$\leq 20$} words and a verb root.
It is possible for multiple sentences to have identical \eat{}s, when they only differ in e.g. contractions or punctuation.
When this happened, we used the \emph{shortest} candidate (in word count) as the surface manifestation of the \eat to monitor the uniqueness of each \eat-sentence mapping.
SNLI had $570155$ sentences and our SNMT-subset $5747035$ sentences overall.

\subsubsection{Approach}
We focused on the following grammatical features of the main clause: \emph{force} (declarative/question), \emph{negation} (affirmed/negated), \emph{voice} (active/passive), \emph{tense} (past/present and perfect), and \emph{aspect} (simple/progressive). Since the main verb always occupies the E-role of the first \eat-triplet (Section \ref{sec:EAT}), we received the grammatical features from there. We separated these features, and represented the rest of the \eat in a string format by concatenating all the remaining elements. We then built a nested dictionary from such representations to their surface manifestations with different main clause grammar, and constructed parallel corpora from those sentences that differed in \emph{only one grammatical feature}.
This technique is computationally efficient, as it only requires a \emph{single run} over the corpus, regardless of how many grammatical classes are considered.

\begin{table}[t]
\begin{center}
\begin{tabular}{|l|l|cc|l|} \hline
\multirow{2}{*}{\textbf{Feature}} & \multirow{2}{*}{\textbf{Value pairs}} & \multicolumn{2}{c|}{\textbf{No. pairs}} & \multirow{2}{*}{\textbf{Example pair (from SNLI)}} \\
&& \textbf{SNLI} & \textbf{SNMT} & \\ \hline
\multirow{2}{*}{Aspect} & simple & \multirow{2}{*}{$5753$} & \multirow{2}{*}{$4436$} & A black dog digs in the snow. \\ & progressive &&& A black dog is digging in the snow. \\ \hline
\multirow{2}{*}{Tense} & present & \multirow{2}{*}{$1061$} & \multirow{2}{*}{$19674$} & The dog runs through the water. \\ & past &&& the dog ran through the water \\ \hline
\multirow{2}{*}{Negation} & affirmed & \multirow{2}{*}{$858$} & \multirow{2}{*}{$15317$} & The man has a musical instrument. \\ & negated &&& The man does not have a musical instrument. \\ \hline
\multirow{2}{*}{Voice} & active & \multirow{2}{*}{$57$} & \multirow{2}{*}{$61$} & A woman is hitting a tennis ball. \\ & passive &&& A tennis ball is being hit by a woman. \\ \hline
\multirow{2}{*}{Tense} & present/past & \multirow{2}{*}{$27$} & \multirow{2}{*}{$3711$} & A man falls down \\ & (plu)perfect &&& A man has fallen down. \\ \hline
\multirow{2}{*}{Force} & declarative & \multirow{2}{*}{$1$} & \multirow{2}{*}{$9648$} & You are coming with me for coffee. \\ & question &&& Are you coming with me for coffee. \\ \hline
\end{tabular}
\caption{Parallel corpora between main clause grammatical features, derived from SNLI \newline
(``Full'': identical except for main clause grammar; ``Main clause'': identical main clause arguments).}
\label{tab:SNLI-parallel-corpora}
\end{center}
\end{table}

\subsubsection{Results}
We generated six parallel corpora from both SNLI and SNMT, listed in Table \ref{tab:SNLI-parallel-corpora} with example sentences and corpus sizes.
The number of pairs varied significantly between classes, e.g. present-past pairs being common but active-passive pairs rare in both corpora.
Nevertheless, our technique was able to construct parallel corpora reliably between pairs of all classes. Manual evaluation on subsets of each corpus confirmed that the pairs were appropriate.

\subsubsection{Summary}
As far as we know, parallel corpora like these have so far not been available in the NLP community.
They have potential use in NLP tasks like NMT-based \emph{style transfer} \citep{Wubbenetal2012, Xuetal2012}.
In addition to the corpora themselves, we provide the algorithms for obtaining similar parallel corpora from any corpus parsed into the PCFG or dependency grammar format.
We hope our contributions will advance research on related topics and extend the range of available methods.
Furthermore, the success of \eat in parallel corpus creation demonstrates its viability for various other \emph{information retrieval} tasks, where it might be useful to e.g. ignore certain grammatical features and only focus on core argument structure. The flexibility and easy manageability of \eat makes it optimal for such tasks.

%
%

\subsection{Generating English from \eat}
\label{sec:EAT2seq}

Generating natural language from a semantic representation like \eat is needed in NLP tasks that require presenting the semantic information as human-readable text.
It can also be applied for \emph{text transformation} by altering the representation prior to generation.
We used an encoder-decoder network to generate English from \eat-input.
We call this technique \emph{\eatseq} (Section \ref{sec:experiment-settings}) and applied it for \emph{sentence reconstruction} and \emph{grammatical transformation} (Section \ref{sec:Sentence-reproduction}).
We additionally devised a rule-based alternative for grammatical transformation, using the \emph{SimpleNLG} surface realizer (Section \ref{sec:experiment-settings}).
Both systems use \eat-input, with the addition of the original dependency-parsed sentence for SimpleNLG.
Moreover, we employed \emph{post-transformation comparison} between the transformation's \eat and the desired target \eat.
This provides a highly effective method for evaluating transformation success. By retaining only successful transformations, we built additional parallel corpora between grammatical classes in $14$ directions.


\subsubsection{Data}
We used the same SNMT dataset as in Section \ref{sec:parallel-corpora}, which we divided to a training set ($4987680$ sentences), a validation set ($4574$ sentences), and a test set ($754781$ sentences).
We vectorized \eat-tuples by using $0/1$ for the Boolean grammatical features, and pre-trained embeddings for words (lemmatized by SpaCy).
As the embeddings we used $300$-dimensional GloVe vectors \citep{Glove} trained on a Common Crawl corpus.\footnote{\url{http://commoncrawl.org/}. The GloVe vectors are included in SpaCy's large English model.} Each \eat-vector thus had $3 \times 300 + 28 = 928$ components, where $28$ is the number of grammatical features (Section \ref{sec:grammatical-features}).
We limited our training and validation sets to sentences that did not contain words unknown to the embedding matrix,
but did not enforce this condition in the test set.

\subsubsection{Approach}
\label{sec:experiment-settings}

\begin{figure}[t]
\centering
\includegraphics[scale=0.22]{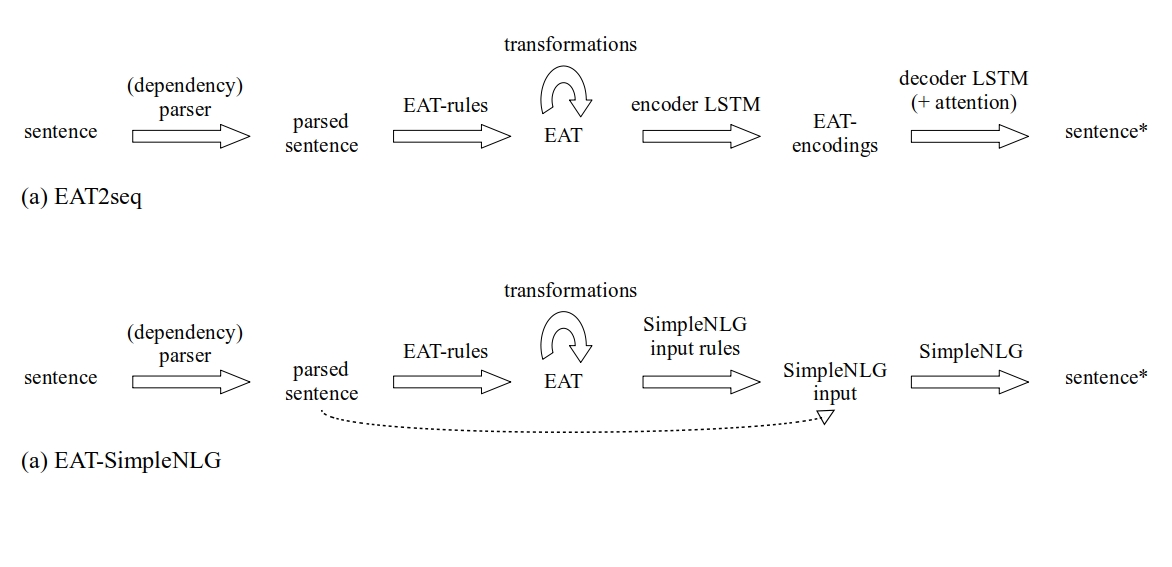}
\caption{\eatseq and \eatsnlg pipelines}
\label{fig:EAT2seq}
\end{figure}

\noindent{\textbf{\eatseq.}}
We implemented \eatseq in Python $3$ with PyTorch,\footnote{\url{https://pytorch.org/}}.
The encoder and decoder were two-layer LSTM networks with $600$ hidden units in each layer.
In the forward pass, we initialized the decoder hidden state with the final encoder output, and applied \textit{attention} by using intermediate encoder outputs as additional decoder inputs \citep{Bahdanauetal14}.
We used the batch size of $64$, the negative log-likelihood loss function, and a dropout probability of $0.1$.
With the Adam optimizer \citep{ADAM14}, we began with a learning rate of $0.001$, reducing it to $0.0001$ at epoch $3$ and $0.00001$ at epoch $7$.
After epoch $9$ validation loss no longer decreased, and we used the weights from this epoch in our experiments.
Figure \ref{fig:EAT2seq}a shows the \eatseq pipeline.

\label{sec:EAT-similarity-reward}
We trained \eatseq with greedy search, but used \emph{beam search} in the test phase (with the beam size $10$).
We produced the \eat of each candidate in the final beam, again using SpaCy for dependency parsing.
If a candidate's \eat was identical to the original, we chose it as our final sentence. If no such candidate existed, we compared the \emph{first vector} in each \eat-sequence to the first vector of the original, and rewarded identical components by adding $1$ to the log-probabilities. The first vector corresponds to the main verb and its arguments (Section \ref{sec:EAT}).

\noindent{\textbf{\eatsnlg.}}
\label{sec:SimpleNLG}
As a rule-based alternative to the encoder-decoder network, we built another \eat-based surface realization system that uses \emph{SimpleNLG} \citep{Gatt:Reiter2009}.\footnote{\url{https://github.com/simplenlg/simplenlg}}
Here, we used the dependency-parsed sentence itself additional input, which allowed us to utilize SimpleNLG's ability of directly including complex strings as modifiers.
%
%
We allocated the main verb and its A- and T-roles to the corresponding argument positions in SimpleNLG. We then added three types of additional modifiers: \emph{front-modifiers} come before any arguments; \emph{pre-modifiers} between the grammatical subject and the verb; and \emph{post-modifiers} after the verb and its arguments. We derived the modifiers from the original parse, and added them as such without analyzing them further in SimpleNLG.
This focus on only the main clause allows the system to be applied for sentences of arbitrary complexity (only limited by SpaCy's parsing abilities).
The \eatsnlg pipeline is shown in Figure \ref{fig:EAT2seq}b.

\noindent{\textbf{Experiments.}}
To evaluate how well \eatseq could retain information in the original sentence, we \emph{reproduced} all sentences in the test set.
In addition, we gathered sentences representing different grammatical classes and applied \emph{grammatical transformations} by changing the relevant grammatical feature in the \eat-input.
We experimented with the following classes:
\textit{force} (declarative, question), \textit{truth} (affirmed, negated), \textit{voice} (active, passive), \textit{tense} (present, past, perfect, pluperfect), and \textit{aspect} (perfective, imperfective).
Tense was the only class with more than two variants, and here we transformed between the present tense and all others.
We applied each transformation to both directions with both \eatseq and \eatsnlg.

\noindent{\textbf{Evaluation.}}
For measuring text reproduction performance we used BLEU, METEOR, and the exact match rate between original and reproduced sentences.
We evaluated grammatical transformation success based on the grammatical features of the transformed sentence's \eat.
The transformed sentence should also retain other content in the original sentence,
but applying similarity metrics directly on the transformed sentence would be inappropriate.
Instead, we adopted the idea of using \textit{back-translation} to evaluate machine translation \citep{Rapp09}:
we re-transformed the transformed sentence back to the original class and then calculated the similarity metrics.

In addition, we checked whether the transformed sentence's \eat was identical with the desired target \eat, i.e. the original sentence's \eat with appropriate changes to the relevant grammatical features. This provides the closest criterion we have for ``perfect'' success (assuming correct parsing by SpaCy).
This is a very useful addition, as it allows only retaining perfect outcomes for e.g. generating synthetic parallel corpora. Here, the perfect success rate predicts the size of the resulting corpus as a fraction of the size of the whole dataset the transformations are applied to.

\begin{table}[t]
\begin{center}
\begin{tabular}{|c|c|c|c|c|c|c|c|}
\hline
\multirow{2}{*}{\textbf{Direction}} & \multirow{2}{*}{\textbf{System}} & \multirow{2}{*}{\shortstack{\textbf{Correct} \\ \textbf{target class}}} & \multirow{2}{*}{\shortstack{\textbf{Correct} \\ \textbf{target \eat}}} & \multicolumn{3}{|c|}{\textbf{Back-transformation similarity}} \\ \cline{5-8}
&&&& \textbf{BLEU} & \textbf{METEOR} & \textbf{Exact match} \\ \hline

\multirow{2}{*}{\shortstack{declarative-question}}
& \eatseq & $71.2\%$ & $31.2\%$ & $49.51$ & $31.17$ & $18.5\%$ \\
& SimpleNLG & $66.6\%$ & $35.9\%$ & $82.25$ & $53.99$ & $31.2\%$ \\ \hline

\multirow{2}{*}{\shortstack{question-declarative}}
& \eatseq & $98.6\%$ & $58.6\%$ & $63.81$ & $40.76$ & $30.0\%$ \\
& SimpleNLG & $94.2\%$ & $69.7\%$ & $78.76$ & $50.25$ & $23.1\%$ \\ \hline

\multirow{2}{*}{\shortstack{affirmed-negated}}
& \eatseq & $88.8\%$ & $39.2\%$ & $54.77$ & $34.28$ & $22.3\%$ \\
& SimpleNLG & $91.0\%$ & $59.9\%$ & $88.79$ & $59.60$ & $47.8\%$ \\ \hline

\multirow{2}{*}{\shortstack{negated-affirmed}}
& \eatseq & $99.2\%$ & $57.6\%$ & $63.43$ & $40.35$ & $24.3\%$ \\
& SimpleNLG & $97.5\%$ & $62.9\%$ & $81.03$ & $50.44$ & $26.0\%$ \\ \hline

\multirow{2}{*}{\shortstack{active-passive}}
& \eatseq & $73.8\%$ & $8.0\%$ & $35.11$ & $24.29$ & $5.1\%$ \\
& SimpleNLG & $88.5\%$ & $37.4\%$ & $82.65$ & $54.22$ & $33.8\%$ \\ \hline

\multirow{2}{*}{\shortstack{passive-active}}
& \eatseq & $92.5\%$ & $23.2\%$ & $48.95$ & $30.09$ & $12.0\%$ \\
& SimpleNLG & $95.5\%$ & $56.6\%$ & $87.98$ & $58.07$ & $45.2\%$ \\ \hline

\multirow{2}{*}{\shortstack{present-past}}
& \eatseq & $93.8\%$ & $47.2\%$ & $56.86$ & $35.70$ & $24.3\%$ \\
& SimpleNLG & $93.3\%$ & $66.5\%$ & $87.91$ & $58.29$ & $44.9\%$ \\ \hline

\multirow{2}{*}{\shortstack{past-present}}
& \eatseq & $85.5\%$ & $39.5\%$ & $52.49$ & $32.86$ & $25.1\%$ \\
& SimpleNLG & $89.3\%$ & $65.6\%$ & $89.71$ & $60.95$ & $54.5\%$ \\ \hline

\multirow{2}{*}{\shortstack{present-perfect}}
& \eatseq & $85.1\%$ & $40.6\%$ & $55.08$ & $34.55$ & $21.2\%$ \\
& SimpleNLG & $92.8\%$ & $65.8\%$ & $87.95$ & $58.14$ & $44.8\%$ \\ \hline

\multirow{2}{*}{\shortstack{perfect-present}}
& \eatseq & $83.2\%$ & $40.6\%$ & $57.13$ & $35.75$ & $19.8\%$ \\
& SimpleNLG & $81.4\%$ & $54.0\%$ & $85.33$ & $55.24$ & $35.9\%$ \\ \hline

\multirow{2}{*}{\shortstack{present-pluperfect}}
& \eatseq & $90.6\%$ & $43.8\%$ & $55.35$ & $34.77$ & $22.3\%$ \\
& SimpleNLG & $91.9\%$ & $65.3\%$ & $87.86$ & $58.05$ & $44.6\%$ \\ \hline

\multirow{2}{*}{\shortstack{pluperfect-present}}
& \eatseq & $84.6\%$ & $33.5\%$ & $52.93$ & $32.52$ & $20.7\%$ \\
& SimpleNLG & $92.0\%$ & $63.7\%$ & $90.17$ & $60.48$ & $52.83\%$ \\ \hline

\multirow{2}{*}{\shortstack{simple-progressive}}
& \eatseq & $72.4\%$ & $32.2\%$ & $53.10$ & $33.32$ & $19.4\%$ \\
& SimpleNLG & $90.5\%$ & $59.1\%$ & $87.87$ & $58.69$ & $45.7\%$ \\ \hline

\multirow{2}{*}{\shortstack{progressive-simple}}
& \eatseq & $95.8\%$ & $42.5\%$ & $53.44$ & $33.86$ & $21.0\%$ \\
& SimpleNLG & $83.5\%$ & $50.4\%$ & $85.43$ & $55.13$ & $35.9\%$ \\ \hline

\end{tabular}
\caption{Grammatical transformation performance of \eatseq and \eatsnlg on $10000$ test sentences ($\leq 20$ words) from each grammatical class. BLEU averaged from $1-4$-grams.}
\label{tab:Grammatical-transformation}
\end{center}
\end{table}

\subsubsection{Results}
\label{sec:experimental-results}


\noindent{\textbf{Sentence reproduction.}}
\label{sec:Sentence-reproduction}
\eatseq reached a sentence reproduction BLEU score of $64.99$ (averaged from \mbox{$1-4$}-grams), a METEOR score of $40.97$, and an exact match rate of $22\%$.
These results indicate that \eatseq successfully retained most information from the original sentences.
For comparison, in the experiments of \citet{Coughlin2003} that compared BLEU to human evaluation on a \mbox{$1-4$} scale from low to high, BLEU over $60$ systematically correlated with a $\geq 3$ grade.

\noindent{\textbf{Grammatical transformation.}}
Table \ref{tab:Grammatical-transformation} shows grammatical transformation performance with both \eatseq and \eatsnlg, and Table \ref{tab:Example-transformations} displays examples.
In Table \ref{tab:Grammatical-transformation}, \emph{correct target class} (column $3$) means that the output sentence's \eat has the intended target class; and \emph{correct target \eat} (column $4$) means that the entire \eat of the output sentence is identical to the input \eat after appropriate transformations to the grammatical features. Both columns show percentages of output sentences that fulfilled the respective criteria. Those output sentences with the correct target \eat can be considered as \emph{perfect successes}, assuming correct parsing by SpaCy.

\eatseq performance evidently correlated with target class frequency in the training set.
Transformations toward minority classes were the most challenging: especially question, passive, and progressive.
In contrast, question-declarative, negated-affirmed, and progressive-simple transformation succeeded $\geq 95\%$ of the time in reaching the target class.
Except in voice transformation, average back-transformation BLEU was between $48-59$, METEOR between $31-41$, and exact match rate between $18\%-30\%$.
In active-passive transformation the transformation had the desired target \eat $8\%$ of the time, in passive-active $23.2\%$ of the time, and in the rest $33\%-59\%$ of the time.

In back-transformation similarity metrics, \eatsnlg results are consistently higher compared to \eatseq.
This is unsurprising, given that everything except main clause argument structure is essentially copy-pasted to the transformation (Section \ref{sec:SimpleNLG}).
\eatsnlg also had better overall success rate, but less consistently.
An exception was declarative-question, which on manual evaluation was discovered to be partly due to SimpleNLG failing to conduct he transformation when the main verb was the copula (\emph{be}). Such dependence on the proper functioning of SimpleNLG is thus a shortcoming of \eatsnlg,
which \eatseq does not suffer from.

To produce novel parallel corpora, we retained only those sentence pairs where the target had the desired \eat.
Manual evaluation on subsets of these corpora confirmed that the pairs were appropriate.
Hence, while success rates varied between classes, this \eat-comparison metric allows effective filtering of successful transformations.
We provide these synthetic parallel corpora as additional contributions.

\begin{table}[t!]
\begin{center}
\begin{tabular}{|c|p{3.3cm}|p{3.25cm}|p{3.25cm}|}
\hline

\multirow{2}{*}{\textbf{Direction}} & \multicolumn{1}{c|}{\multirow{2}{*}{\textbf{Original}}} & \multicolumn{2}{c|}{\textbf{Transformation}} \\ \cline{3-4}
&& \multicolumn{1}{c|}{\textbf{\eatseq}} & \multicolumn{1}{c|}{\textbf{\eatsnlg}} \\ \hline

\multirow{2}{*}{declarative-question} &
" maybe they just left town . " &
maybe did they just leave town ? &
did they just leave town " maybe ? \\ \hline
 
question-declarative &
'don't you want it ?' &
you do n't want it . &
' you do not want it . \\ \hline

affirmed-negated &
he's a genie . &
he 's not a genie . &
he is not a genie . \\ \hline

negated-affirmed &
it's not my brother's . &
it 's my brother . &
it is my brother 's . \\ \hline

\multirow{2}{*}{active-passive} &
\multirow{2}{*}{i need my own life .} &
my own life is needed by my . &
my own life is needed by me . \\ \hline

\multirow{2}{*}{passive-active} &
we weren't scanned by that probe . &
\multirow{2}{*}{that probe did n't scan us .} &
\multirow{2}{*}{that probe did not scan us .} \\ \hline

\multirow{2}{*}{present-past} &
the mother basks in her triumph . &
the mother squirmed in her triumph . &
the mother basked in her triumph . \\ \hline

past-present &
well , she said so once . &
well , she says so so . &
well, she says so once . \\ \hline

present-perfect &
she doesn't have gas . &
she has n't had gas . &
she has not had gas . \\ \hline

\multirow{2}{*}{\shortstack{present-pluperfect}} &
don't you need to have evidence ? &
had n't you needed to have evidence ? &
had you not needed to have evidence ? \\ \hline

\multirow{2}{*}{\shortstack{pluperfect-present}} &
more than half of them had once worked for her . &
more than half of them work work for her . &
more than half of them once works for her . \\ \hline

simple-progressive &
ok , i did . &
i was doing , ok . &
ok , i was doing. \\ \hline

progressive-simple &
haven't you been listening ? &
have n't you listened ? &
have you not listened ? \\ \hline

\end{tabular}
\caption{Example transformations by \eatseq and \eatsnlg}
\label{tab:Example-transformations}
\end{center}
\end{table}

\subsubsection{Summary}
The relatively high text reproduction performance of \eatseq illustrates that \eat retains most information from the original sentence, and presents it in a format usable for an encoder-decoder LSTM network.
We further demonstrated \eatseq's and \eatsnlg's applicability for grammatical transformation.
Since this task has not been focused on in recent NLP research, our results provide strong baselines for future work.
Post-transformation comparison of the output sentence's \eat with the input \eat further makes it possible to retain only successful target sentences, ensuring the appropriateness of the output for e.g. parallel corpus generation.
\section{Discussion}
\label{sec:discussion}

We presented \eat: a novel semantic representation format for NLP that directly builds on theoretical ideas behind the \emph{conjunctivist} framework \citep{Pietroski05a, Pietroski18}.
While \eat bears partial resemblance to prior formats, its main novelty is its \emph{simplicity}.
Using only three semantic roles with positional encoding, it avoids explicit semantic metapredicates and allows simple vectorization via concatenating word embeddings in their respective positions.
By adding grammatical features and presenting multiple \eat-tuples in sequence (with conjunctive default interpretation), we efficiently represent most of the sentence's content with a bare minimum of structure.

EAT's linearity and the lack of metapredicates makes it easy to navigate and use as direct input to e.g. an encoder-decoder network.
This \emph{versatility} is another of its main benefits compared to alternative semantic representation formats like AMR or MRS.
In particular, positional encoding of semantic roles is only practically feasible with a small number of roles, which neither AMR or MRS has.
While this does not mean that \eat should replace them in all use cases, its minimal structure brings about unique benefits especially for flexibility across NLP tasks.

The current variant of \eat implements the basic version of Pietroski's conjunctivist system (Section \ref{sec:conjunctivism}).
Beyond that, additional semantic types and combinatory mechanisms would be needed to account for various non-conjunctive aspects of semantics, such as operators, non-subsective modifiers, connectives, quantifiers, and clausal arguments (Section \ref{sec:conjunctivism-refinements}).
These are currently assimilated to simpler conjunctive variants: operators and non-subsective modifiers are treated simply as normal modifiers; connectives are assimilated to prepositions with clausal arguments; and clausal T-roles are not separated between Theme and Content interpretations.
As discussed in Section \ref{sec:EAT-refinements}, including such information to \eat would be easy simply by adding markers of first- vs. second-order interpretation and Theme vs. Content interpretation to the Boolean features. Our reason for not implementing this was that the information is not available in the syntactic parse alone.
A comparable issue thus arises with \emph{any} semantic parsing framework that abstains from using external lexical information, and is not a problem for \eat as such.

We applied \eat to three NLP tasks: \emph{parallel corpus extraction} between grammatical classes, \emph{text reconstruction} from \eat, and \emph{grammatical transformation}.
To our knowledge, our parallel corpora are so far unique in kind.
Our technique can also be applied to any parsed corpus to extract similar parallel corpora.
Using an external parser (e.g. SpaCy) further allows extracting them from \emph{any} English corpus.
Beyond parallel corpus extraction, \eat can be helpful in other information retrieval tasks involving semantic content, grammar, or their combination. Given the ease of navigating \eat due to its simple structure, such possibilities are open-ended.

The encoder-decoder based \eatseq network was largely able to reconstruct English from \eat.
We demonstrated EAT2seq's applicability for grammatical transformation, along with a rule-based alternative using SimpleNLG.
Since this task has not been focused on in recent NLP research, our results provide strong baselines to compare against in future work.

\section{Related work}
\label{sec:Related-work}

\eatseq combines symbolic methods with deep neural network architectures.
Such \emph{hybrid} approaches have been applied in prior work on \textit{natural language understanding} \citep{Garretteetal11, Lewis:Steedman13, Beltagyetal16}.
\eatseq adopts a standard NMT architecture \citep{Luongetal2015, GoogleTranslate2016}, with the exception of using \eat-sequences as encoder inputs.
\eatseq separates between grammatical and thematic information, allowing for controlled transformation. Partially similar ideas have been presented in prior work on \textit{style transfer} \citep{Sennrichetal16, Rao:Tetreault18, Shenetal2017, Fuetal2018, Shettyetal2018} and \textit{controlled text generation} \citep{Huetal17, Juutietal2018}.
AMR \citep{Banarescuetal13} and MRS \citep{Copestakeetal2005} have been translated to English relying on parallel corpora \citep{Castro-Ferreiraetal17, Gildeaetal18, Cohnetal18, Hajdiketal2019}. On the other hand, Neo-Davidsonian logical forms have been produced directly from dependency parses \citep{Reddyetal2016, Reddyetal2017}. We combine these lines of work by first constructing the \eat-representation from the syntactic parse, and then reconstructing English from \eat.
%
%

Outside highly task-specific rule-based approaches \citep{Ahmed:Lin14, Biluetal15, Baptista16}, grammatical transformation has not been a focus in contemporary NLP.
\citet{Logeswaranetal2018} used a \emph{generative adversarial network} \citep{Goodfellowetal2014} for transforming textual properties, including grammatical class. Since they did not report any semantic evaluation results, comparison is impossible without access to their model. However, example transformations they provide illustrate that their approach can conflate grammatical and lexical properties: e.g. treating \emph{live} as the active variant of \emph{(be) born}.

Rule-based solutions have been proposed for specific tasks like sentence negation \citep{Ahmed:Lin14, Biluetal15}, style transfer \citep{Khosmood:Levinson10, Khosmood12}, and active-passive transformation \citep{Baptista16}.
Compared to such traditional approaches, our transformation rules are maximally simple: changing a single Boolean feature per grammatical class.
Uniquely, both \eatseq and \eatsnlg are applicable \emph{without} the need for elaborate transformation rules or task-specific training for different target classes.
%
\section{Conclusions and future work}
\label{sec:Conclusions}

\eat is the first generic NLP application of the conjunctivist framework in semantics.
Despite drastically reducing the number of semantic roles from alternative formats, it maintains expressiveness by optimizing these roles motivated by the theory.
The versatility of \eat makes its possible uses in information retrieval tasks open-ended.
\eatseq could also be applied to other forms of text transformation beyond changing grammatical class, such as altering lexical arguments themselves.
Further theoretical work connecting \eat more systematically to different semantic frameworks (like DRT) as well as alternative linguistic analyses would also be important extensions.

In line with much prior work in NMT \citep{Luongetal2015, GoogleTranslate2016}, we used LSTMs with attention in the encoder-decoder network.
More recently, the \emph{Transformer} architecture has demonstrated strong performance in sequence-to-sequence mapping \citep{Vaswanietal2017}.
An important further task is thus to optimize \eat-input for classifier architectures beyond LSTMs, especially Transformers.
Another crucial future application is applying \eat beyond English.
The tripartite role scheme abstracts away from surface word order, and would allow variation between inflectional systems.
Modifications would thus mostly be confined to the grammatical features.
There is also theoretical work suggesting the linguistic framework we used (Section \ref{sec:argument-structure-syntax}) to be applicable across languages, at least on an appropriate level of abstraction (e.g. \citealt{Ramchand:Svenonius2014, Wiltschko14}). While this remains controversial, it is worth exploring how well \eat and its linking to syntax could be extended to multiple languages with minimal changes.

\medskip
\small
\noindent \textbf{Acknowledgements.}
We thank prof. N. Asokan and Dr. Mika Juuti for valuable discussions related to the project, and Luca Pajola for helping with early-stage implementation.
Tommi Gr\"{o}ndahl was funded by the Helsinki Doctoral Education Network in Information and Communications Technology (HICT).

\medskip
\small
\noindent \textbf{Competing interests.}
The author(s) declare none.

\bibliography{LF-refs}
\bibliographystyle{nlelike}

\end{document}